\documentclass{article}
\usepackage{arxiv_techreport}

\usepackage{amsmath,amsfonts,bm}

\def\eqref#1{equation~\ref{#1}}
\def\1{\bm{1}}

\DeclareMathAlphabet{\mathsfit}{\encodingdefault}{\sfdefault}{m}{sl}
\SetMathAlphabet{\mathsfit}{bold}{\encodingdefault}{\sfdefault}{bx}{n}

\usepackage{amsmath}
\usepackage{booktabs}
\usepackage{graphicx}
\usepackage{float}
\usepackage{wrapfig}
\usepackage[section]{placeins}
\usepackage{fontawesome5}
\usepackage{subcaption}

\NewDocumentCommand{\runzhe} { mO{} }{\textcolor{blue}{\textsuperscript{\textit{runzhe:}}\textsf{\textbf{[#1]}}}}

\definecolor{hanGreen}{RGB}{0,100,60}
\NewDocumentCommand{\han} { mO{} }{\textcolor{hanGreen}{\textsuperscript{\textit{han:}}\textsf{\textbf{[#1]}}}}

\title{EvoPolicyGym: Evaluating Autonomous Policy Evolution in Interactive Environments}

\author{%
\begin{minipage}[t]{0.88\textwidth}
\centering
Zhilin Wang\textsuperscript{1,*}, Han Song\textsuperscript{2,*},
Runzhe Zhan\textsuperscript{3,*}\\
Jusen Du\textsuperscript{4}, Jiacheng Chen\textsuperscript{2},
Tianle Li\textsuperscript{2}, Qingyu Yin\textsuperscript{5},
Yulun Wu\textsuperscript{5}\\
Zhennan Shen\textsuperscript{9}, Tong Zhu\textsuperscript{6},
Yanshu Li\textsuperscript{7}, Guanjie Chen\textsuperscript{9} \\
Derek F. Wong\textsuperscript{3}, Yafu Li\textsuperscript{2,\textdagger},
Yu Cheng\textsuperscript{2,\textdagger}, Yang Yang\textsuperscript{9,\textdagger}\\[0.35em]
{\normalfont\small 
\textsuperscript{1}University of Science and Technology of China \quad
\textsuperscript{2}The Chinese University of Hong Kong \\
\textsuperscript{3}University of Macau \quad
\textsuperscript{4}Tsinghua University \quad
\textsuperscript{5}Zhejiang University \\
\textsuperscript{6}Soochow University \quad
\textsuperscript{7}Brown University \\
\textsuperscript{9}Shanghai Jiao Tong University \\
} 
{\normalfont\small \textsuperscript{*}Equal contribution \quad
\textsuperscript{\textdagger}Corresponding authors}
\end{minipage}
}

\newcommand{\evopolicygym}{\textsc{EvoPolicyGym}}

\begin{document}

\maketitle

\begin{abstract}
Autonomous agents are increasingly expected to improve executable policies through feedback, yet existing evaluations often collapse this process into a final score or confound it with open-ended software-engineering progress. 
We introduce Autonomous Policy Evolution, a controlled evaluation setting in which a harness–model agent repeatedly edits an executable policy system under a fixed interaction budget. 
We instantiate this setting in EvoPolicyGym, a benchmark built from compact interactive RL environments that evaluates how agents iteratively improve explored policies.
On the EvoPolicyGym suite, GPT-5.5 achieves the strongest aggregate rank score and top-two performance on all 16 environments. 
Beyond leaderboard results, EvoPolicyGym also provides trajectory-level diagnostics that distinguish how agents allocate budget, convert feedback into parametric tuning. 
These analyses show that strong autonomous policy evolution depends not only on isolated task wins, but on discovering task-appropriate mechanisms and refining policies under bounded feedback.

\begin{center}
\small
\faGithub~\href{https://github.com/Linzwcs/EvoPolicyGym}{\texttt{EvoPolicyGym}}
\qquad
\faDatabase~\href{https://huggingface.co/datasets/linzw/EvoPolicyGym-Exp-data}{\texttt{EvoPolicyGym-Exp-data}}\qquad
\faGlobe~\href{https://linzwcs.github.io/EvoPolicyGym/}{\texttt{EvoPolicyGym.io}}
\end{center}
\end{abstract}

\section{Introduction}
Autonomous agents are expected to improve through feedback rather than produce a single fixed output. Modern coding agents are able to call tools, observe failures, and revise executable artifacts over long horizons \citep{sweagent,openhands,codeact}, while self-improvement systems show that language models can use reflection to refine answers across attempts \citep{reflexion,selfrefine,alphaevolve}. However, this broader capability is hard to evaluate because improvement is both an outcome and a process: final scores can hide blind retries, overfitting to visible feedback, brittle special cases, missing verification, and other trajectory-level failure modes \citep{agentlens,codeagentbehavior,howcodingagentsfail}. 
Fully open-ended engineering tasks add further confounders, including evolving specifications and software-maintenance quality \citep{sweci,roadmapbench,specbench,slopcodebench}. 
% We therefore need an evaluation setting that is controlled enough to isolate primitive self-evolution abilities, while still forcing agents to improve an executable behavior through real environment feedback.
We therefore need a controlled setting that isolates an agent's ability to convert bounded environment feedback into generalizable improvements of an executable policy, while retaining the iterative decisions that make autonomous improvement difficult.

We address this gap by formalizing \emph{Autonomous Policy Evolution}, a problem in which an agent repeatedly revises an executable decision policy using feedback from prior deployments. 
Formally, the observable object is the sequence of submitted policy systems and train-feedback records, whereas the outcome is the held-out return of the checkpoint selected on hidden validation.
The goal is not only to maximize observed performance alone, but to produce a policy that generalizes to environment instances after limited interaction. 
The bounded budget is therefore part of the capability being measured: it requires each agent to choose what information to acquire, when to explore or exploit, and how efficiently to convert sparse behavioral evidence into robust policy improvement.

To evaluate this problem, we instantiate \emph{Autonomous Policy Evolution} in \textsc{EvoPolicyGym}, a controlled benchmark built from compact interactive environments. 
Unlike open-ended engineering benchmarks \citep{frontiereng}, 
\textsc{EvoPolicyGym} makes policy evolution itself the evaluated object: an agent repeatedly edits an executable policy system, submits it under a fixed interaction budget, and receives server-generated feedback from sandboxed rollouts. 
Train submissions return aggregate and trajectory-level feedback, whereas validation and held-out cases remain server-side.
This protocol makes policy evolution itself the evaluated object, rather than direct task execution or open-ended engineering progress. Across environments derived from standard reinforcement-learning substrates \citep{mujoco,minigrid,gymnasium}, \textsc{EvoPolicyGym} records the full execution--feedback--revise trajectory, enabling comparisons of not only final held-out performance but also how agents diagnose failures, allocate budget, and balance exploration with exploitation.

We conduct preliminary experiments on the classical RL gymnasium environment.
We evaluate four harness--model agents on Core16, a 16-environment suite spanning Gym/Box2D, MuJoCo, MiniGrid, and robotics/driving tasks, under a common 128-episode interaction budget. The results show that GPT-5.5 obtains the highest aggregate rank score and top-two performance on all 16 environments, whereas Claude Opus 4.7 leads the MiniGrid family. The remaining agents achieve isolated task wins but substantially lower cross-environment coverage.
% The results show that policy-evolution performance is highly environment-dependent: 

% Beyond these scores, our trajectory analysis shows that agents adopt different feedback-use strategies, including patching observed failures, using visual perception for control refinement, and trading off early exploration against late exploitation.
Beyond final scores, our analysis on aggregate edit statistics and selected-policy structure reveal systematic differences between structural synthesis and parameter tuning, while audited CarRacing and BipedalWalker traces illustrate how individual agents translate visible feedback into revisions.
Thus, \textsc{EvoPolicyGym} serves both as a leaderboard and as a diagnostic substrate for studying how self-evolving agents interact with environment feedback under bounded budgets.

To summarize, our contributions are threefold:
\begin{itemize}
    \item We formulate \emph{Autonomous Policy Evolution} as a benchmarkable setting for evaluating agents in policy searching tasks.
    \item We instantiate this setting in \textsc{EvoPolicyGym}, a controlled benchmark with strict visibility boundaries, bounded interaction, trajectory-level feedback, and hidden held-out generalization. 
    \item We introduce trajectory-level diagnostics that relate policy improvement to budget-conditioned policy evolution and audited trace revisions.
\end{itemize}

\section{Related Work}
\paragraph{From static patches to long-horizon coding-agent evaluation.}
% Repository-level benchmarks such as SWE-bench evaluate coding agents on
% execution-grounded software-engineering tasks, requiring them to edit real
% repositories and validate patches against project-level tests \citep{swebench}.
% Agent systems built around this setting further emphasize repository navigation,
% tool use, test execution, and agent-computer interfaces in realistic workflows
% \citep{sweagent,openhands,codeact}.

Repository-level benchmarks such as SWE-bench evaluate coding agents on execution-grounded software-engineering tasks, where agents edit real repositories and validate patches against unit tests \citep{swebench}. 
Systems built on this setting emphasize repository navigation, tool use, and interactive debugging in realistic workflows \citep{sweagent,codeact,openhands}. Yet one-shot patch generation only partially captures software development, where code must be revised, extended, and maintained over time. Long-horizon benchmarks therefore move beyond single-edit success to study software evolution under repeated modifications, revealing issues such as quality degradation, multi-file consistency challenges, and mismatches between visible validation and hidden behavioral tests \citep{slopcodebench,sweci,sweevo,roadmapbench,specbench}. In contrast to these long-horizon benchmarks driven by evolving specifications and discrete unit-test outcomes, we study \emph{policy evolution under bounded feedback}, where agents iteratively refine executable policies from limited rollout signals and are evaluated by continuous return rather than binary test outcomes.

% \runzhe{above text summarizes the literature but does not yet state our technical distinction. i suggest adding a closing contrast like, ... ``In contrast to benchmarks driven by coding tests, \evopolicygym{} holds the task contract fixed and evaluates how an agent improves a persistent decision policy from bounded rollout feedback, with xxxxx measurement.''}

\paragraph{Feedback-driven self-improvement.}
A line of research has shown that language models can improve their performance through iterative feedback rather than single-shot generation. Reflexion and Self-Refine leverage language-level reflection to refine outputs across attempts \citep{reflexion,selfrefine}. Building on this idea, Voyager, Eureka, FunSearch, and AlphaEvolve extend self-improvement to executable artifacts, enabling iterative improvement of skills, reward functions, programs, and algorithms via external feedback, thereby broadening the scope of self-evolution from language outputs to algorithmic and programmatic representations \citep{voyager,eureka,funsearch,alphaevolve}. Extending this further, our work studies this paradigm at the level of \emph{harness--model agents}, focusing on the ability of agents to iteratively improve their \emph{interaction-driven policies} through bounded environment feedback in a unified execution loop.

\paragraph{Evaluation of interactive and self-improving agents.}
Feedback-driven approaches have led to a range of interactive benchmarks that place agents in web, operating-system, database, and workplace environments to assess tool use, state tracking, and multi-step decision making \citep{agentbench,webarena,osworld,workarena}. However, these settings are primarily episodic, focusing on task completion within single interactions, and do not capture how agents iteratively improve a persistent policy across repeated deployments. A related line of experimentation benchmarks studies iterative machine-learning system development, where agents design, execute, and refine models or training pipelines under repeated feedback \citep{mlagentbench,mlebench}. While these settings introduce longer optimization loops, they remain centered on task-specific system construction rather than general policy evolution across a unified interaction interface.

\paragraph{Bounded optimization and trajectory-level analysis for agents.}
Frontier-Eng studies generative engineering design under bounded optimization, where agents improve executable artifacts under explicit limits on feedback and computation\citep{frontiereng}. It covers a broad class of engineering environments, whereas we consider standard reinforcement learning environments and study how agents iteratively improve decision-making policies through environment interaction. In this setting, interaction budgets are structured at the episode level, and feedback is mediated through a platform-controlled evaluation protocol that enforces consistent interaction constraints across episodes. These design choices make the setting suitable for trajectory-level analysis. Prior work has shown that aggregate success rates can obscure intermediate behaviors such as retry patterns, failure recovery loops, and verification-related issues \citep{agentlens,codeagentbehavior,swechat,howcodingagentsfail}. The episode-level budgeting and controlled feedback mechanism provide a more consistent and comparable basis for analysis, enabling finer-grained attribution of agent behavior along interaction trajectories.

\section{\evopolicygym{}: A Framework for Autonomous Policy Evolution}
\label{sec:framework}

\begin{figure*}[t]
    \centering
    \includegraphics[width=0.98\textwidth]{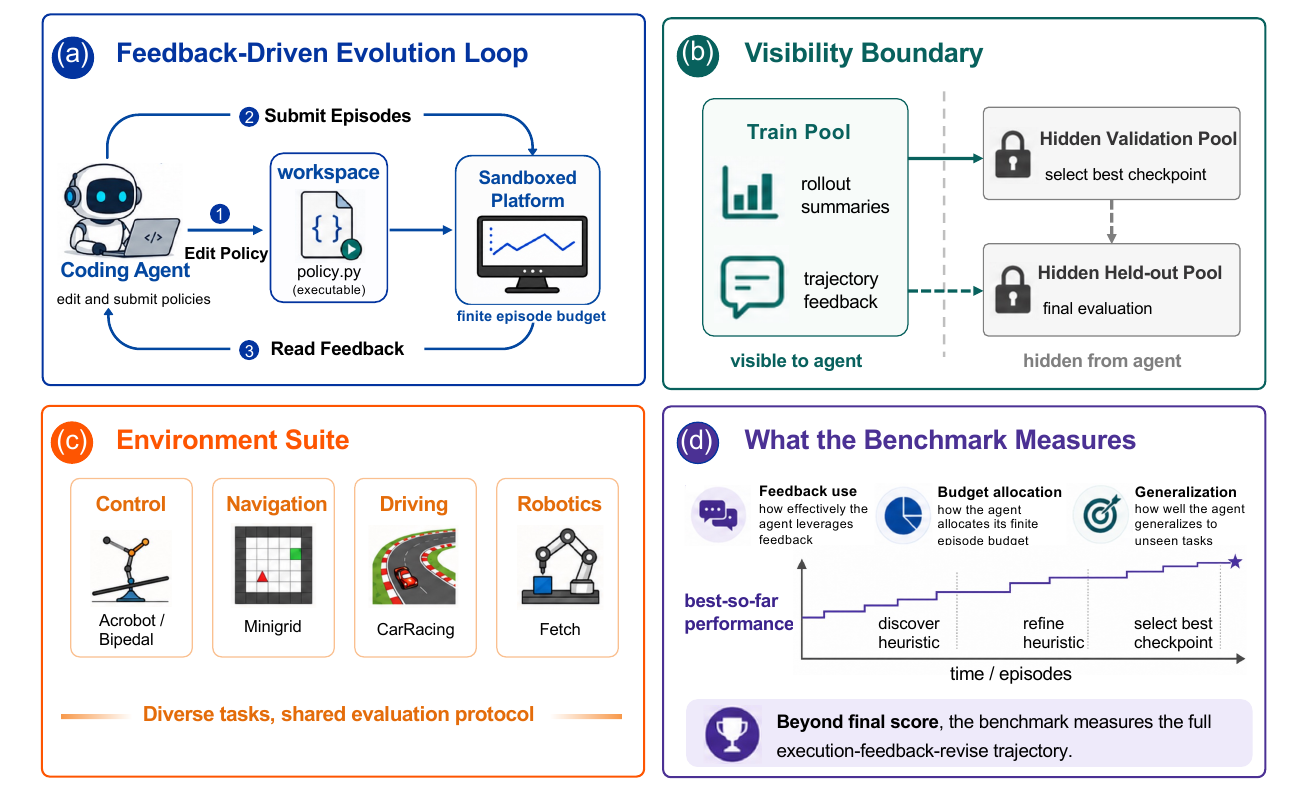}
    \caption{\small
    \textbf{EvoPolicyGym framework.} (a) \textbf{Interaction loop}: agents edit policies, submit episodic rollouts under a finite budget, and receive platform-mediated feedback. 
    (b) \textbf{Visibility boundary}: training feedback is visible, while validation-based checkpoint selection and held-out evaluation are hidden.
    (c) \textbf{Environment suite}: a unified interface spanning control, navigation, driving, and robotics tasks under a shared evaluation protocol. 
    (d) \textbf{Measured aspects}: feedback utilization, budget efficiency, and policy improvement dynamics, captured via the evolution of best-so-far performance over time.
    }
    % \caption{
    %     Main \evopolicygym{} online optimization protocol.
    %     The agent sees the workspace-facing interfaces and self-evolves by
    %     editing \texttt{system/}, submitting train episodes, and reading
    %     \texttt{feedback/}. Server-owned finalization remains hidden:
    %     validation selects the best strategy and heldout evaluation writes the
    %     final score to \texttt{run.json}.
    % }
    \label{fig:online-submit-protocol}
\end{figure*}

\evopolicygym{} frames autonomous policy evolution as an agent-driven
optimization loop for executable decision policies. In each run, a coding agent
maintains a persistent policy workspace, submits candidate revisions for visible
train episodes, receives server-generated rollout summaries and trajectories,
and revises the policy under a fixed episode budget. 
The primary evaluation unit is a complete budget-constrained run, scored by the held-out return of its best validation checkpoint. The associated trajectory provides diagnostic evidence about how that outcome was reached.
Figure~\ref{fig:online-submit-protocol} summarizes the visible workspace
interfaces, the submission loop, and server-owned finalization.
%\runzhe{this phrase conflicts with the scoring rule, which uses the held-out return of one validation-selected checkpoint. from my understanding, trajectories support diagnosis rather than leaderboard scoring}
%\runzhe{suggested edit: The primary evaluation unit is a complete budget-constrained run, scored by the held-out return of its best validation checkpoint. The associated trajectory provides diagnostic evidence about how that outcome was reached.}

\subsection{Environment, Policy, and Interaction}
\label{sec:task-objects}

Before specifying the benchmark protocol, we define the three objects that make
a policy-evolution task executable: an environment, a policy system, and an
episode. The definitions use the common \texttt{reset}/\texttt{step}
environment interface \citep{gymnasium}, but are stated in benchmark-facing
terms because 
%\evopolicygym{} evaluates coding agents \runzhe{these days, everyone is calling it a ``coding agent'', right?} that optimize
executable decision policies.

\paragraph{Environment.}
An \emph{environment} is an interactive task in which a policy acts. It exposes
observations, accepts actions, and returns rewards that measure task progress.
A reset initializes an episode; each later \texttt{step(action)} advances the
environment and returns the next observation, reward, and termination status.
Observations may be pixels, vectors, dictionaries, or symbolic grids, and
actions may be discrete choices or continuous controls.

\paragraph{Policy.}
A \emph{policy} is the decision rule that chooses actions using observations
and any information it has retained from earlier interaction. We write this
memory as an internal state $h_t$. A deterministic policy maps
$(o_t,h_t)$ to an action and updated internal state,
$(a_t,h_{t+1})=\mu(o_t,h_t)$. A stochastic policy analogously samples
$(a_t,h_{t+1}) \sim \pi(\cdot \mid o_t,h_t)$. In \evopolicygym{}, the
stateful mapping is implemented by the executable Python artifact that the coding agent writes and the judge later runs. Its minimal judge-facing interface is an object with \texttt{reset} at the start of an episode and \texttt{act(obs)} at each step; the internal state is maintained behind this interface. The artifact may also include helper modules, constants, planners, controllers, diagnostics, or learned parameters. We call this whole executable bundle the \emph{policy system}.
%\runzhe{concerns regarding the policy notation... too restrictive here. a stateful policy is not generally a function of $o_t$ alone because our experimental environment explicitly permits memory. formally, define an internal state $h_t$ here, so that we can derive $(a_t,h_{t+1})=\mu(o_t,h_t)$, with the stochastic analogue if needed. then, reserve ``policy system'' for the executable bundle that implements this stateful mapping}

\begin{figure}[t]
\centering
\begin{minipage}[t]{0.485\linewidth}
\textbf{Gymnasium-style env (server-owned)}
\begin{tcblisting}{
  enhanced,
  listing only,
  equal height group=env-policy-api,
  width=\linewidth,
  colback=cardbg,
  colframe=accent,
  boxrule=0.35pt,
  arc=2pt,
  left=6pt,
  right=6pt,
  top=4pt,
  bottom=4pt,
  boxsep=0pt,
  listing options={
    language=Python,
    basicstyle=\ttfamily\scriptsize,
    keywordstyle=\color{accent}\bfseries,
    commentstyle=\color{black!55},
    stringstyle=\color{black!70},
    emph={Environment,reset,step},
    emphstyle=\color{accent}\bfseries,
    columns=fullflexible,
    keepspaces=true,
    showstringspaces=false,
    breaklines=true,
    breakatwhitespace=true
  }
}
class Environment:
    observation_space = ...
    action_space = ...

    def reset(self, *, seed=None,
              options=None): ...
    # -> obs, info

    def step(self, action): ...
    # -> obs, reward, terminated,
    #    truncated, info
\end{tcblisting}
\end{minipage}
\hfill
\begin{minipage}[t]{0.485\linewidth}
\textbf{Policy-system entry point}
\begin{tcblisting}{
  enhanced,
  listing only,
  equal height group=env-policy-api,
  width=\linewidth,
  colback=cardbg,
  colframe=accent,
  boxrule=0.35pt,
  arc=2pt,
  left=6pt,
  right=6pt,
  top=4pt,
  bottom=4pt,
  boxsep=0pt,
  listing options={
    language=Python,
    basicstyle=\ttfamily\scriptsize,
    keywordstyle=\color{accent}\bfseries,
    commentstyle=\color{black!55},
    stringstyle=\color{black!70},
    emph={Policy,__init__,reset,act},
    emphstyle=\color{accent}\bfseries,
    columns=fullflexible,
    keepspaces=true,
    showstringspaces=false,
    breaklines=true,
    breakatwhitespace=true
  }
}
class Policy:
    def __init__(self, obs_space,
                 action_space, env_meta): ...

    def reset(self, episode_index): ...
    # -> None

    def act(self, obs): ...
    # -> action


    
\end{tcblisting}
\end{minipage}
\caption{Minimal runtime boundary for one \evopolicygym{} episode. The server
owns the Gymnasium-style environment, which maps actions to observations and
rewards. The submitted policy system maps observations to actions through this
entry point, but may internally contain helper modules, planners, memory,
diagnostics, learned parameters, or other decision logic.}
\label{fig:environment-policy-api}
\end{figure}

\paragraph{Episode.}
An \emph{episode} is one complete policy--environment interaction. It begins
with an environment reset and ends at termination or truncation. At time $t$,
the policy receives observation $o_t$, returns action $a_t$, and the environment
produces reward $r_t$ and the next observation. The episode is the basic
evaluation unit in \evopolicygym{}; its \emph{return} is the cumulative reward
collected during the interaction, and higher return is better in our benchmark.

\subsection{Autonomous Policy Evolution Protocol}
\label{sec:ape}

\evopolicygym{} evaluates the ability of a coding agent to optimize an
executable policy system from environment feedback. A run begins with one
environment, an initial policy workspace, and a fixed episode budget. Over the
run, the agent inspects the workspace, reads feedback from previous
submissions, edits code, and decides when to submit the current artifact and
how much of the remaining episode budget to spend on evaluation.

At observed revision $i$, let $W_i$ denote the workspace state visible to the
agent, $F_i$ the server-written feedback from prior submissions, and $B_i$ the
remaining episode budget. The current executable policy system is induced by
the workspace, written $P_i=\Phi(W_i)$. We denote the coding agent by
$\pi_\theta$, a language model together with its tool-using harness. The agent
observes $(W_i,F_i,B_i)$, carries history $\mathcal{H}_i$, and writes patches
to the workspace.
Figure~\ref{fig:online-submit-protocol} organizes the same loop around the
\emph{Agent}, \emph{Workspace}, and \emph{Server}.

The run induces a sequence of observed-state transitions:
\[
\pi_\theta(W_i, F_i, B_i, \mathcal{H}_i)
\rightarrow (u_i,s_i,\mathcal{H}_{i+1}),\qquad
s_i\in\{\bot\}\cup\mathcal{C}(B_i),
\]
\[
W_{i+1}=\mathrm{apply}(W_i,u_i),\qquad
P_{i+1}=\Phi(W_{i+1}),
\]
\[
(\Delta F_i,c_i)=S(B_i,P_{i+1},s_i),\qquad
B_{i+1}=B_i-c_i,\qquad
F_{i+1}=F_i\cup\Delta F_i.
\]
Here $\mathcal{H}_i$ is the agent's accumulated conversational and tool-use
history, $u_i$ is a workspace patch, and $s_i$ is a server-facing submit
command. The null command $s_i=\bot$ means that no train evaluation is requested
at this revision; otherwise $s_i\in\mathcal{C}(B_i)$ specifies a valid train
submit under the remaining episode budget. A patch may tune constants, add
helper modules, introduce memory, replace a controller, add diagnostics, or
restructure the policy system induced by the workspace. The server operator $S$
returns the new feedback $\Delta F_i$ and charged episode count $c_i$. It sets
$c_i=0$ and returns no new feedback when $s_i=\bot$ or no evaluation is
accepted; otherwise it snapshots the selected revision, charges the accepted
train episodes, and returns feedback for later patches. Submit commands do not
themselves modify $W_i$ or $P_i$. After the run ends, the server
automatically performs hidden validation selection and held-out evaluation.
Final scores therefore compare each agent's optimization outcome under the same episode budget.

\FloatBarrier
\subsection{Feedback and Evaluation Boundary}
\label{sec:feedback-signals}

The agent obtains environment evidence only through submitted train episodes.
For each accepted submit, the server evaluates a snapshot of the current policy
system, charges one budget unit per requested episode, and writes the
observable feedback signal $F_i$. Feedback includes structured summaries,
episode returns and statuses, trajectory records, diagnostic streams, error
reports, and optional environment-specific artifacts such as frames or videos.
These signals guide the next update $u_i$ to $P_i$.

A fixed visibility boundary separates online feedback from final evaluation.
Train episodes provide in-loop rollout evidence; validation and held-out
evidence remain hidden until the optimization loop ends.
%validation score -> post-hoc diagnostics in Section 4.3
%\runzhe{Also call the score evolution curves in Section~\ref{sec:score-evolution}: post-hoc diagnostics wherever they first appear, so readers cannot mistake hidden validation for an optimization signal.} 
The server then
selects a checkpoint by hidden validation and reports its hidden held-out
performance.
Appendix~\ref{app:evaluation-and-scoring} gives the detailed visibility table,
scoring rule, and audit traces.

\subsection{Environment Abstraction and Extensibility}
\label{sec:gym-compatible-envs}

The environment layer is separated from the agent protocol. Any
Gymnasium-compatible episodic environment with a \texttt{reset}/\texttt{step}
interface can be wrapped by an adapter that implements reset, step, and
reproducible episode initialization. The adapter converts observations and
actions into compact schemas, loads reproducible train/validation/held-out
initialization splits, and keeps hidden split metadata outside the agent
workspace.

This design keeps interaction, visibility, budget, and artifact semantics fixed
as environment coverage grows. We distinguish adapter-level support from full
experimental validation. At the adapter level, the current implementation
includes Gymnasium-style wrappers and interface tests for Classic Control, Toy
Text, Box2D, MuJoCo, Atari/ALE, MiniGrid, MiniWorld, HighwayEnv,
Gymnasium-Robotics, MO-Gymnasium, BrowserGym, MiniWoB++, and MetaWorld. These
tests check that tasks can be reset, stepped, initialized from reproducible
splits, and exposed through the common observation/action schema. At the
experimental level, this paper validates the full protocol on the calibrated
Core16 subset, which uses 16 tasks spanning Gym/Box2D, MuJoCo, MiniGrid, and
robotics/driving environments. The broader adapter surface serves as a task
reservoir from which future experiments can select additional calibrated
subsets.
%\runzhe{Distinguish implemented and validated support from packages merely available in the environment. Or weaken the claim to: ``The adapter design is compatible with these Gymnasium-style families; the present study validates the protocol on the xxxxx subset.''}

\section{Experiments}
\label{sec:experiments}

% \subsection{Core16 Suite and Agents}
% \label{sec:core16-suite}

% We evaluate four harness--model agents on the Core16 suite listed in
% Appendix~\ref{app:core16-suite-details}. The suite covers four environment
% families: Gym / Box2D, MuJoCo, MiniGrid, and Robotics / Driving. Each run gives
% the agent a 128-episode interaction budget. The agent decides how to spend this
% budget across candidate policy evaluations and visible train cases. After the
% optimization loop ends, hidden validation selects the checkpoint from that run,
% and hidden held-out evaluation measures final generalization. We use 16
% validation cases and 32 held-out cases per environment. Tables also include a
% uniform random-policy reference evaluated on the same held-out pools; it is not
% a training agent and receives no interaction budget.

% The leaderboard entries are complete harness--model agents. GPT-5.5 is run
% through the Codex harness; Claude Opus 4.7, MiniMax-M3, and DeepSeek-V4-Pro are
% run through the Claude Code-compatible harness. All agents face the same suite,
% budget, validation size, and held-out size. We report validation-selected
% held-out mean return for agents and held-out mean return for the random-policy
% reference; higher is better within each environment.
% Cross-environment summaries use the rank-score defined in
% Section~\ref{sec:scoring-audit}. Conventional RL baselines are outside this
% first leaderboard because their training interface differs from the interactive
% code-editing setting studied here.

\subsection{Experimental Setup}
\label{sec:experimental-setup}

We instantiate the evaluation framework on the Core16 suite listed in
Appendix~\ref{app:core16-suite-details}. Core16 covers four environment families: Gym / Box2D, MuJoCo, MiniGrid, and Robotics / Driving. For each environment, all agents receive a 128-episode training budget, with 16 validation cases and 32 held-out cases reserved for server-side selection and final evaluation. We evaluate each model together with its coding harness and include the harness as part of the evaluation dimension: GPT-5.5 is run through the Codex harness, while Claude Opus 4.7, MiniMax-M3, and DeepSeek-V4-Pro are run through the Claude Code harness. All agents face the same Core16 environment suite, case splits, submission interface, and scoring protocol. 

We do not normalize token use, context management, or provider-specific inference defaults across harnesses; these are part of the evaluated model-and-harness system, and token statistics are reported only as diagnostics in Appendix~\ref{app:token-accounting}. Each leaderboard cell is one 128-episode optimization run for one model together with its coding harness on one environment, followed by hidden-validation checkpoint selection and held-out evaluation. We report validation-selected held-out mean return for each agent, with higher values indicating better performance within each environment. A uniform random policy is also evaluated on the same held-out pools.
Conventional RL baselines are outside this leaderboard because their training interface differs from the interactive code-editing setting studied here. The 128-episode budget is also far below the sample regime in which standard RL training methods are expected to converge; giving them the much larger budgets needed for convergence would change the comparison being measured.
%\runzhe{exact token use, budget, and inference settings. zhilin: In coding agent setting, temperature. max token length, etc. are not emphasized. }. 
%\runzhe{if there is only one 128-episode optimization run per cell, say so plainly}
% Cross-environment summaries
% use the rank-score defined in Section~\ref{sec:scoring-audit}. 

\subsection{Leaderboard Results}
\label{sec:leaderboard-results}
\label{sec:aggregate-ranking}

Tables~\ref{tab:core16-raw} and~\ref{tab:aggregate} give two complementary readings of the same Core16 experiments. Table~\ref{tab:core16-raw} reports the validation-selected held-out return in each environment, preserving the native reward scale for within-environment comparison. Because these reward scales are not comparable across tasks, Table~\ref{tab:aggregate} summarizes each model together with its coding harness using the rank score defined in Appendix~\ref{app:evaluation-and-scoring}. The aggregate score should therefore be read as a reliability measure across heterogeneous tasks, not as an average return.

\begin{table}[t]
\caption{Core16 final held-out scores for each environment. Each cell is the validation-selected policy's mean return on that environment's held-out episodes, reported on the environment's native reward scale; values are comparable within an environment but not across environments. The uniform random-policy reference is evaluated on the same held-out pools. Best values are bolded, second-best values are underlined, and rows are sorted by Core16 aggregate rank score.}
\label{tab:core16-raw}
\begin{center}
\scriptsize
\setlength{\tabcolsep}{3pt}
\textbf{A. Gym / Box2D and MuJoCo}\par\vspace{0.25em}
\resizebox{\textwidth}{!}{%
\begin{tabular}{llrrrrrrrr}
\toprule
 & & \multicolumn{4}{c}{Gym / Box2D $\uparrow$} & \multicolumn{4}{c}{MuJoCo $\uparrow$} \\
\cmidrule(lr){3-6}\cmidrule(lr){7-10}
Model & Harness & Acrobot & Cont.Car & Bipedal & CarRacing
& Reacher & HalfCheetah & Ant & Pusher \\
\midrule
GPT-5.5 & Codex & \textbf{-84.688} & \underline{95.216} & \textbf{248.874} & \textbf{604.145} & \textbf{-3.473} & \underline{601.833} & \underline{989.580} & \textbf{-37.106} \\
Claude Opus 4.7 & Claude Code & \underline{-88.594} & \textbf{98.774} & \underline{-15.844} & \underline{602.337} & \underline{-3.979} & 452.022 & \textbf{990.149} & \underline{-38.520} \\
MiniMax-M3 & Claude Code & -136.406 & 91.881 & -80.874 & 233.435 & -5.103 & \textbf{606.154} & 983.425 & -39.245 \\
DeepSeek-V4-Pro & Claude Code & -91.156 & 94.477 & -97.475 & 25.199 & -6.506 & -0.468 & 897.995 & -54.180 \\
Random policy & Uniform & -499.156 & -33.359 & -100.976 & -29.627 & -43.772 & -291.844 & -34.602 & -147.462 \\
\bottomrule
\end{tabular}
}

\vspace{0.75em}
\textbf{B. MiniGrid and Robotics / Driving}\par\vspace{0.25em}
\resizebox{\textwidth}{!}{%
\begin{tabular}{llrrrrrrrr}
\toprule
 & & \multicolumn{4}{c}{MiniGrid 
 $\uparrow$} & \multicolumn{4}{c}{Robotics / Driving $\uparrow$} \\
\cmidrule(lr){3-6}\cmidrule(lr){7-10}
Model & Harness & DoorKey & KeyCorr. & FourRooms & Obst.Maze
& Parking & Roundabout & FetchPush & FetchPick \\
\midrule
GPT-5.5 & Codex & \textbf{0.986} & \underline{0.918} & \underline{0.664} & \underline{0.904} & \textbf{-30.237} & \underline{9.818} & \textbf{-18.156} & \textbf{-11.812} \\
Claude Opus 4.7 & Claude Code & \underline{0.983} & \textbf{0.930} & \textbf{0.701} & \textbf{0.911} & -38.181 & 9.331 & \underline{-19.688} & -13.344 \\
MiniMax-M3 & Claude Code & 0.000 & 0.000 & 0.403 & 0.000 & \underline{-32.710} & 9.521 & -25.875 & \underline{-13.219} \\
DeepSeek-V4-Pro & Claude Code & 0.000 & 0.000 & 0.079 & 0.000 & -53.384 & \textbf{10.203} & -27.656 & -20.562 \\
Random policy & Uniform & 0.000 & 0.000 & 0.042 & 0.000 & -47.448 & 7.188 & -46.031 & -43.750 \\
\bottomrule
\end{tabular}

}
\end{center}
\end{table}

\begin{table}[t]
\caption{Aggregate Core16 leaderboard. Family and Core16 scores are macro-averages of per-environment rank scores over the four agents and the uniform random-policy reference. \emph{Wins} counts first-place environments, \emph{Top-2} counts first- or second-place environments, and rows are sorted by Core16 score.}
\label{tab:aggregate}
\begin{center}
\scriptsize
\setlength{\tabcolsep}{3.5pt}
\begin{tabular}{llrrrrrrrr}
\toprule
Model & Harness & Gym/Box2D & MuJoCo & MiniGrid & Robotics/Driving
& Core16 & Budget & Wins & Top-2 \\
\midrule
GPT-5.5 & Codex & \textbf{0.938} & \textbf{0.875} & \underline{0.812} & \textbf{0.938} & \textbf{0.891} & 128 & 9 & 16 \\
Claude Opus 4.7 & Claude Code & \underline{0.812} & \underline{0.750} & \textbf{0.938} & 0.500 & \underline{0.750} & 128 & 5 & 12 \\
MiniMax-M3 & Claude Code & 0.375 & 0.625 & 0.500 & \underline{0.625} & 0.531 & 128 & 1 & 3 \\
DeepSeek-V4-Pro & Claude Code & 0.375 & 0.250 & 0.438 & 0.375 & 0.359 & 128 & 1 & 1 \\
Random policy & - & 0.000 & 0.000 & 0.375 & 0.062 & 0.109 & -- & 0 & 0 \\
\bottomrule
\end{tabular}
\end{center}
\end{table}

\paragraph{The leading entries are defined by coverage.}
GPT-5.5 obtains the highest Core16 score ($0.891$), with nine wins and top-two placement on all 16 environments. 
Claude Opus 4.7 ranks second ($0.750$), with five wins and 12 top-two placements. 
The gap is therefore not only a count of first-place finishes: GPT-5.5 is the only entry that remains near the top across every environment, while Claude Opus 4.7 remains second overall through strong coverage and the best MiniGrid family score ($0.938$).

\paragraph{Different task families favor different agents.}
The raw held-out returns explain how these aggregate scores arise. GPT-5.5
leads the Gym / Box2D, MuJoCo, and Robotics / Driving family scores, whereas
Claude Opus 4.7 is strongest on MiniGrid. At the task level, Claude Opus 4.7
wins ContinuousCar, Ant, KeyCorridor, FourRooms, and ObstructedMaze, while
GPT-5.5 supplies the broader set of wins across the remaining families. This
pattern makes the leaderboard more informative than a single global winner: it
shows both the overall reliability ordering and the task families where that
ordering changes.

\paragraph{Local wins do not imply suite-level reliability.}
MiniMax-M3 and DeepSeek-V4-Pro each win one environment, but their aggregate scores remain substantially lower. 
MiniMax-M3 wins HalfCheetah and reaches the top two on Parking and FetchPickAndPlace, yet its weaker Gym / Box2D and MiniGrid ranks reduce its Core16 score to $0.531$. 
DeepSeek-V4-Pro wins Roundabout but has only one top-two placement overall, giving a Core16 score of
$0.359$. 
The uniform random policy scores $0.109$, mainly from shared rank credit on MiniGrid zero-score ties. The leaderboard therefore rewards consistent near-top performance across the suite rather than isolated task success.

\subsection{Post-Hoc Score Trajectories}
\label{sec:score-evolution}

\begin{figure}[t]
    \centering
    \includegraphics[width=0.98\textwidth]{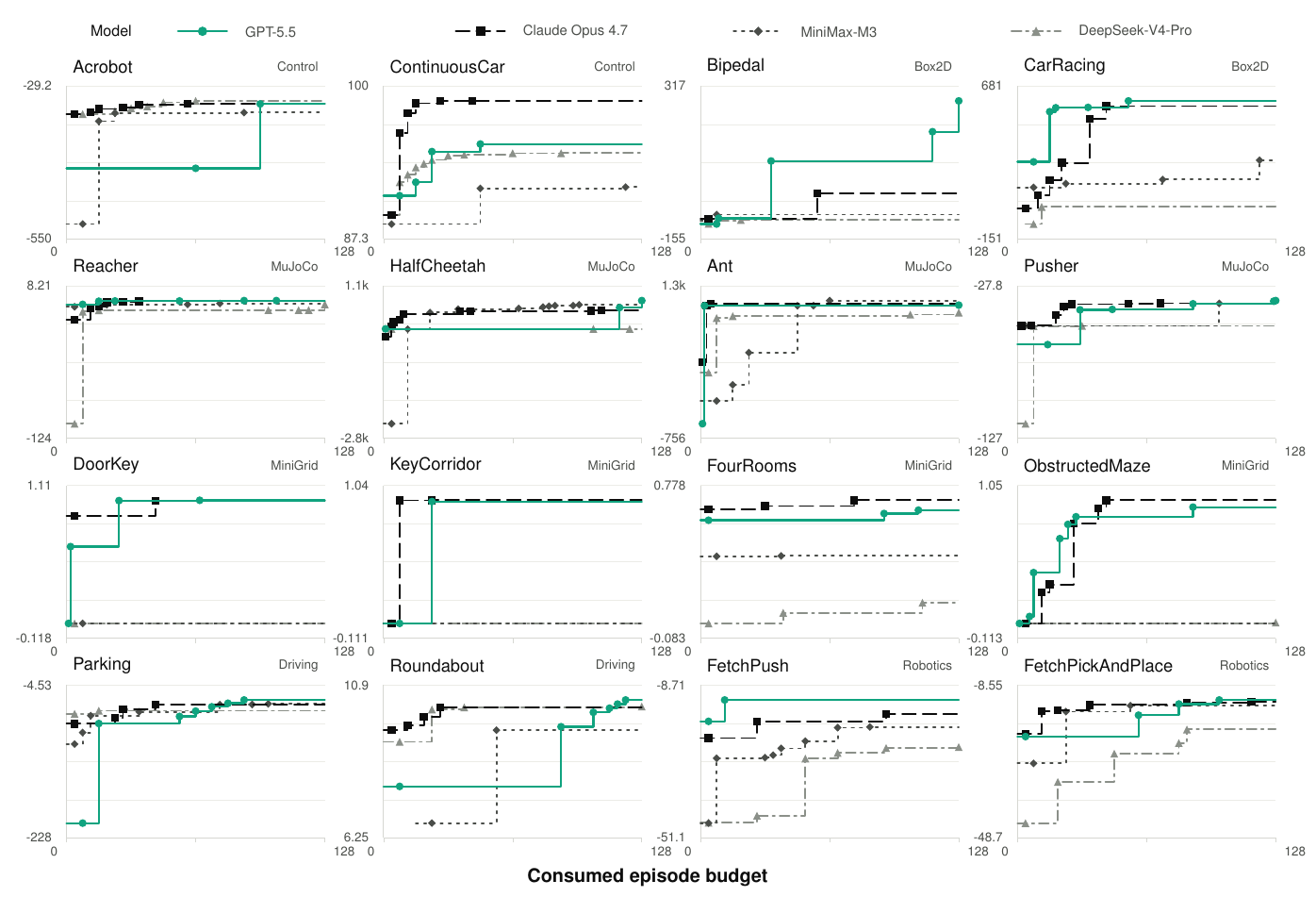}
    \caption{\small Score evolution over each run's episode-budget trajectory. Each curve tracks the post-hoc best-so-far hidden-validation score across candidate evaluations. Vertical jumps indicate improvements in the selected policy, while plateaus correspond to budget spent without improvement.}
    \label{fig:score-evolution}
\end{figure}

The leaderboard aggregates each run into a single selected checkpoint, which
hides the temporal structure of how high-performing policies are discovered.
We therefore compute a post-hoc diagnostic: the evolution of the best-so-far
hidden-validation score over the consumed episode budget
(Figure~\ref{fig:score-evolution}). The agent never observes these
hidden-validation curves during optimization; they are reconstructed after the
run to show when useful candidate policies appeared. These trajectories capture
when improvements occur, how frequently they arise, and how efficiently budget
is converted into better candidate policies.

A key signal in these curves is the occurrence of improvement events over the
budget. Each vertical jump corresponds to the discovery of a higher-quality
candidate policy, while flat segments indicate periods in which additional
budget does not yield improvements on hidden validation. The timing of these
jumps provides a post-hoc efficiency diagnostic, distinguishing agents that
identify high-scoring candidate policies early from agents that consume more
budget before reaching comparable hidden-validation scores.
In Figure~\ref{fig:score-evolution}, MiniGrid panels often show sparse but
sharp jumps, MuJoCo panels show more incremental gains, and several robotics
or driving panels show delayed improvements after substantial budget has been
spent. These curves are computed only after each run is finished; they are used
for analysis, not as feedback that agents can observe during optimization.

These trajectory-level patterns complement the leaderboard results by
distinguishing final performance from the path by which candidates appeared.
They show that similar selected-checkpoint scores can arise from early jumps
followed by plateaus or from late improvements after much of the budget has
already been consumed. Section~\ref{sec:behavior-analysis} provides
qualitative evidence about these trajectories through interaction logs and
code-edit diagnostics, including how agents use feedback, revise code, and preserve selected checkpoints.

%\runzhe{``many''?What specific environments are you referring to? Model-specific timing claims should be accompanied by summary statistics across all 16 trajectories; otherwise restrict them to the environments shown in the figure}
%runzhe{case studies illustrate plausible mechanisms but do not ``explain'' agent-level score differences causally. use ``provides qualitative evidence about'' instead.}

\section{Mechanisms of Policy Evolution}
\label{sec:behavior-analysis}

\subsection{Structural Synthesis and Parametric Tuning}
\label{sec:two-aspects}

The leaderboard tells us which agents win, and
Figure~\ref{fig:score-evolution} shows when their scores improve. We now examine how agents explore the policy implementation space during this process: do they introduce new control mechanisms, or do they refine constants and thresholds within an already plausible controller?

We refer to these two exploration modes as \emph{structural synthesis} and
\emph{parametric tuning}. Structural synthesis creates task machinery such as
perception, memory, planning, reward interpretation, or state abstraction.
Parametric tuning adjusts gains, thresholds, constants, and branch-local
parameters inside a plausible controller. This split follows from the policy
system itself: each submitted policy combines computational structure
(e.g., modules, state, branches, and control logic) with parameters that determine
how that structure behaves. A run can therefore contain both forms of
exploration, so we use the split as a diagnostic lens rather than an exclusive
taxonomy.

We first split environments by what a good policy must contain. The
synthesis-dominant group contains pixel-perception and symbolic-planning tasks,
where a policy must build task-specific machinery such as visual state
extraction, memory, search, or recovery logic. The tuning-dominant group
contains lower-dimensional control tasks, where a simple controller family
often exists and improvement comes mainly from adjusting gains, thresholds, and
branch-local constants. This split lets us ask whether the observed performance
gap comes from building the right control machinery or from tuning a controller
that is already in the right family.

\begin{table}[ht]
\caption{Realized computational structure in validation-selected policies.
Values are group means over the same synthesis/tuning split as
Figure~\ref{fig:two-abilities}. Columns report deterministic AST features of
the selected policy source bundle (\texttt{policy.py} plus reachable local
Python modules).}
\label{tab:code-structure}
\begin{center}
\footnotesize
\setlength{\tabcolsep}{3.5pt}
\begin{tabular}{@{}lllrrrrr@{}}
\toprule
Demand & Agent & Harness & Funcs & Branches & Loops & Depth & State \\
\midrule
Synthesis & GPT-5.5 & Codex & 30.2 & 68.2 & 13.0 & 4.6 & 48.4 \\
Synthesis & Claude Opus 4.7 & Claude Code & 19.0 & 77.0 & 16.8 & 7.6 & 26.0 \\
Synthesis & MiniMax-M3 & Claude Code & 12.8 & 38.2 & 7.6 & 4.0 & 16.0 \\
Synthesis & DeepSeek-V4-Pro & Claude Code & 5.4 & 21.8 & 3.2 & 3.2 & 11.2 \\
\midrule
Tuning & GPT-5.5 & Codex & 8.5 & 8.6 & 0.5 & 2.1 & 9.2 \\
Tuning & Claude Opus 4.7 & Claude Code & 6.5 & 4.5 & 0.5 & 1.7 & 8.2 \\
Tuning & MiniMax-M3 & Claude Code & 4.5 & 7.7 & 0.4 & 2.3 & 6.9 \\
Tuning & DeepSeek-V4-Pro & Claude Code & 3.6 & 5.8 & 0.1 & 2.5 & 5.7 \\
\bottomrule
\end{tabular}
\end{center}
\end{table}

\paragraph{Synthesis tasks need richer machinery.}
Table~\ref{tab:code-structure} compares the validation-selected policy bundles
under this split. It reports deterministic AST features of \texttt{policy.py}
plus self-written local modules reachable by imports. The synthesis-dominant
rows are visibly heavier: the strongest agents select substantially richer
source bundles, with more functions, branches, loops, and persistent state,
while the tuning-dominant rows are much smaller and more compressed.
These features are diagnostic rather than sufficient. 
Nontrivial code volume is therefore not sufficient for strong performance. In other words, complex code is not necessarily a task-adapted mechanism for solving the problem.

%\runzhe{richer code does not cause better performance or capture the right ``task-relevant abstraction''. Replace the final two sentences with: ``Nontrivial code volume is therefore not sufficient for strong performance. In other words, the results point in that direction, but they do not conclusively show that successful synthesis comes from having the right computational structure.'' }

\begin{figure}[!t]
    \centering
    \includegraphics[width=0.98\textwidth]{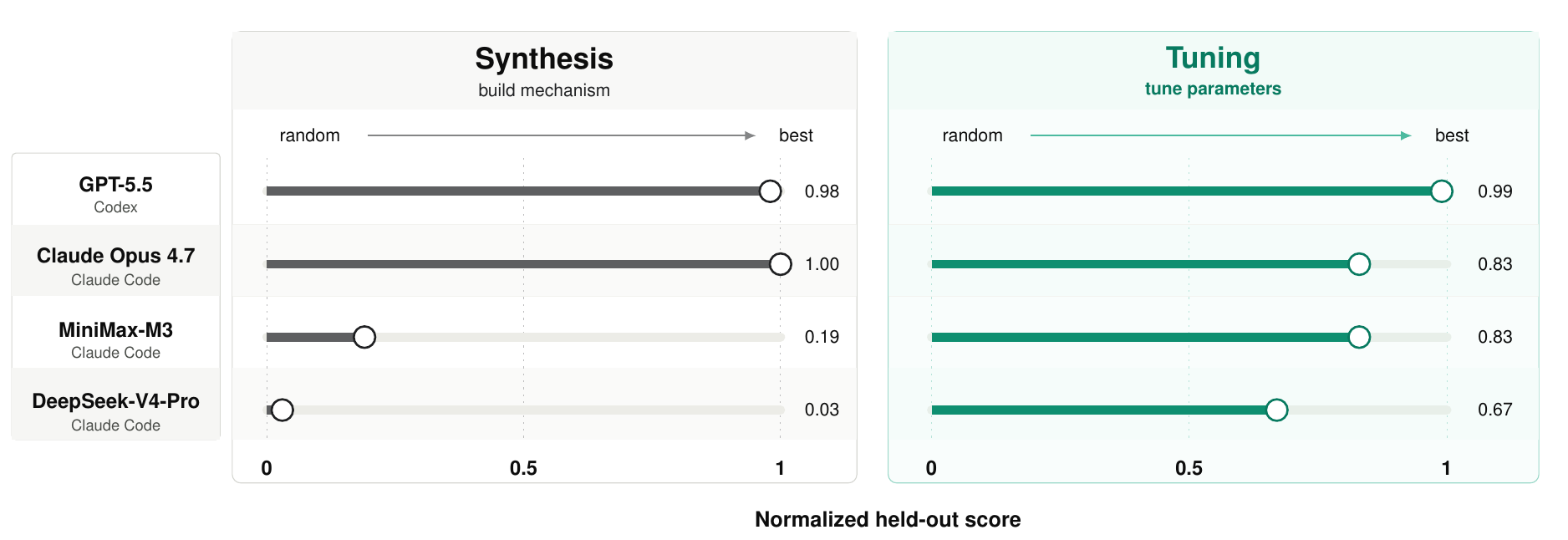}
    \caption{\small Relative held-out performance under different dominant task
    demands. Scores are macro means over synthesis-dominant and
    tuning-dominant environments, using the random-to-best normalization
    described in the text. Each row follows the same model and harness identity
    as the main leaderboard. Per-environment values are reported in Appendix
    Table~\ref{tab:two-abilities-detail}.}
    \label{fig:two-abilities}
\end{figure}

\paragraph{The score gap opens on synthesis tasks.}
Because raw rewards are not comparable across environments, we normalize
held-out scores on a per-environment random-to-best scale before aggregating
the two demand groups in Figure~\ref{fig:two-abilities}:
\[
    \mathrm{norm}_{m,e} =
    \mathrm{clip}_{[0,1]}
    \left(
    \frac{R_{m,e}^{\mathrm{heldout}} - R_{e}^{\mathrm{random}}}
         {R_{e}^{\mathrm{best}} - R_{e}^{\mathrm{random}}}
    \right),
\]
where $R_{e}^{\mathrm{random}}$ is the random-policy anchor on the same
held-out pool and $R_{e}^{\mathrm{best}}$ is the best held-out score achieved
by any evaluated agent on environment $e$. We macro-average
$\mathrm{norm}_{m,e}$ within the synthesis- and tuning-dominant groups. This
diagnostic scale is separate from the rank-based leaderboard score.

%\begin{figure}[H]
%    \centering
%    \includegraphics[width=0.98\textwidth]
%    {figs/two_abilities_domains.pdf}
    %\caption{Ability-level \runzhe{``demand'' or ``ability''? the terms should be unified throughout the main text \& caption...} normalized held-out scores. Scores are macro means
    %over synthesis-dominant and tuning-dominant environments, using the
    %random-to-best normalization described in the text. Each row follows the
    %same model and harness identity as the main leaderboard; per-environment
    %values are reported in Appendix Table~\ref{tab:two-%abilities-detail}.}
%    \label{fig:two-abilities}
%\end{figure}

Figure~\ref{fig:two-abilities} shows that agents separate most on the
synthesis-dominant side. GPT-5.5 and Claude Opus 4.7 nearly reach the best
observed held-out policies ($0.98$ and $1.00$), whereas MiniMax-M3 and
DeepSeek-V4-Pro remain close to the random anchor ($0.19$ and $0.03$) and solve
none of the three locked-door MiniGrid tasks. On tuning-dominant environments,
the same agents cluster more tightly ($0.67$--$0.99$). This is not simply the
leaderboard repeated in two columns: MiniMax-M3 is competitive on tuning
($0.83$), Claude Opus 4.7 misses BipedalWalker ($0.24$), and DeepSeek-V4-Pro
falls below random on Parking. Together with the code and edit diagnostics below, this performance split motivates two candidate failure modes: failing to discover an effective structure and failing to refine a plausible structure.
%\runzhe{final return does not reveal which internal step failed. Suggested wording: ``Together with the code and edit diagnostics below, this performance split motivates two candidate failure modes: failing to discover an effective structure and failing to refine a plausible structure.''}

\paragraph{The winning edit type changes with the task.}
We classify each score-bearing submit transition before measuring whether it
helps. A \emph{synthesis edit} introduces a new policy-bundle AST topology
after numeric constants are stripped. A \emph{parametric edit} changes the
source bundle while preserving that stripped topology. We exclude revisited
topologies and byte-identical retests. %\runzhe{excluding them changes the denominator and may preferentially remove failed candidate actions?report how many transitions are excluded by reason.} 
Table~\ref{tab:edit-types}  then asks which edit type produces new validation bests, counting an edit as a hit only
if it raises the validation best-so-far.

\begin{table}[H]
\caption{Synthesis-edit and parametric-edit success by task category. Each score-bearing submit
transition compares the submitted policy source bundle (\texttt{policy.py} and
reachable local Python modules). A synthesis edit introduces a previously
unseen AST topology after numeric constants are stripped; a parametric edit
changes the source bundle while preserving the immediately previous stripped
topology. Rollback topologies and byte-identical retests are excluded. An edit succeeds if it raises the validation best-so-far.}
\label{tab:edit-types}
\begin{center}
\small
\setlength{\tabcolsep}{4.8pt}
\begin{tabular}{lllrrrr}
\toprule
 & & & \multicolumn{2}{c}{Synthesis edits} & \multicolumn{2}{c}{Parametric edits} \\
\cmidrule(lr){4-5}\cmidrule(lr){6-7}
Agent & Harness & Task & $n$ & hit & $n$ & hit \\
\midrule
GPT-5.5 & Codex & Synthesis & 37 & 41\% & 1 & 100\% \\
 &  & Tuning & 48 & 38\% & 31 & {61\%} \\
\addlinespace
Claude Opus 4.7 & Claude Code & Synthesis & 31 & {48\%} & -- & -- \\
 &  & Tuning & 123 & 26\% & 58 & {21\%} \\
\addlinespace
MiniMax-M3 & Claude Code & Synthesis & 39 & {10\%} & -- & -- \\
 &  & Tuning & 100 & 25\% & 57 & {25\%} \\
\addlinespace
DeepSeek-V4-Pro & Claude Code & Synthesis & 132 & {3\%} & -- & -- \\
 &  & Tuning & 112 & 30\% & 21 & {38\%} \\
\bottomrule
\end{tabular}
\end{center}
\end{table}

The table shows where those improvements come from. On synthesis-dominant
tasks, GPT-5.5 and Claude Opus 4.7 turn synthesis edits into new validation
bests at high rates ($41\%$ and $48\%$), while MiniMax-M3 and
DeepSeek-V4-Pro mostly churn structure without traction ($10\%$ and $3\%$).
Same-topology edits rarely rescue a wrong mechanism, but they become useful on
tuning-dominant tasks once the controller family is close enough.
%\runzhe{table4: check bold logic.}

\subsection{Trajectory Case Studies}
\label{sec:carracing-case}

\begin{figure}[t]
    \centering
    \includegraphics[width=0.98\textwidth]{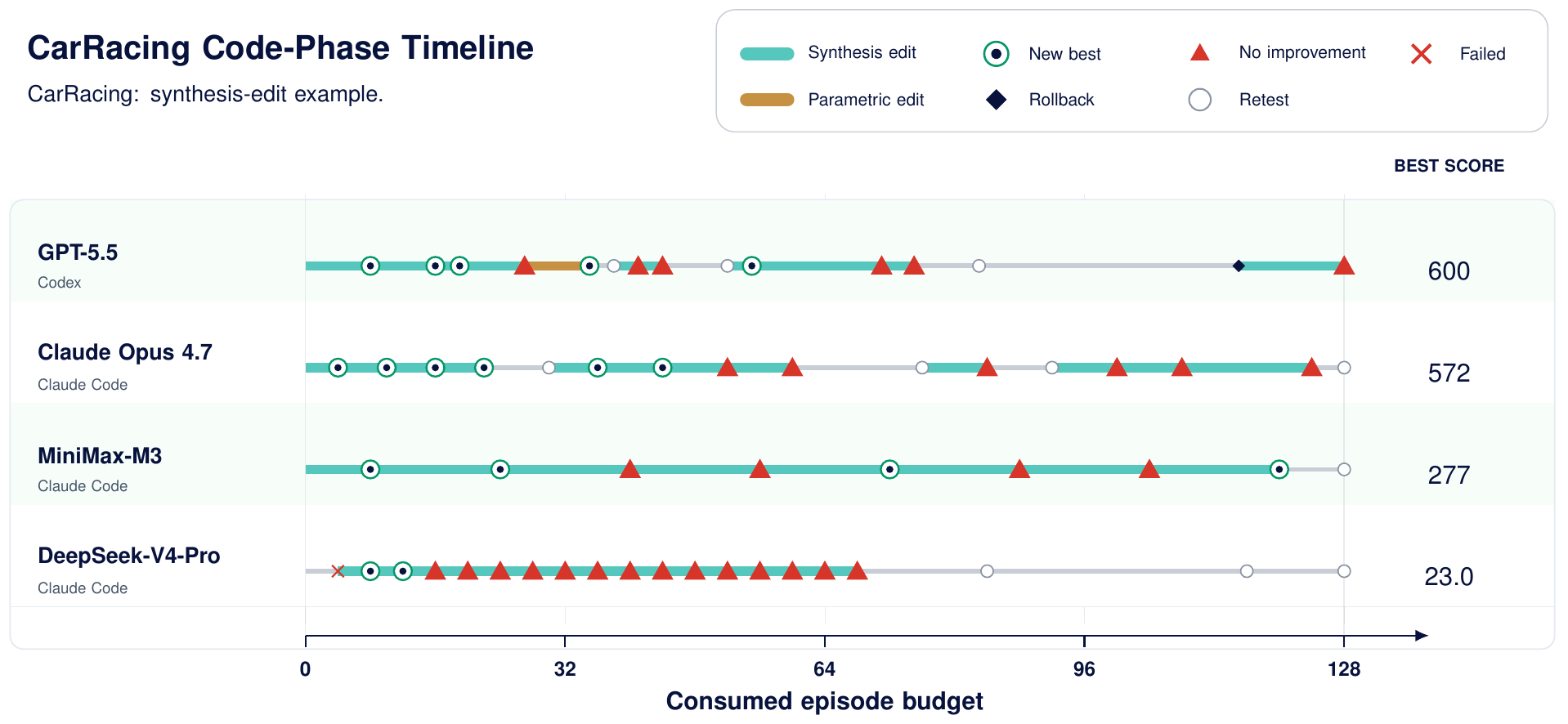}
    \caption{\small CarRacing code-phase timeline. Phase bands are inferred
    mechanically from the same policy source-bundle rule as Table~\ref{tab:edit-types}:
    synthesis-edit phases denote new AST topologies after numeric constants are
    stripped, and parametric-edit phases denote changed source bundles under the
    same topology. Symbols mark
    validation outcomes and candidate-management events; rollback/retest are
    event types, not additional edit types.}
    \label{fig:carracing-code-phase}
\end{figure}

\begin{figure}[t]
    \centering
    \includegraphics[width=0.98\textwidth]{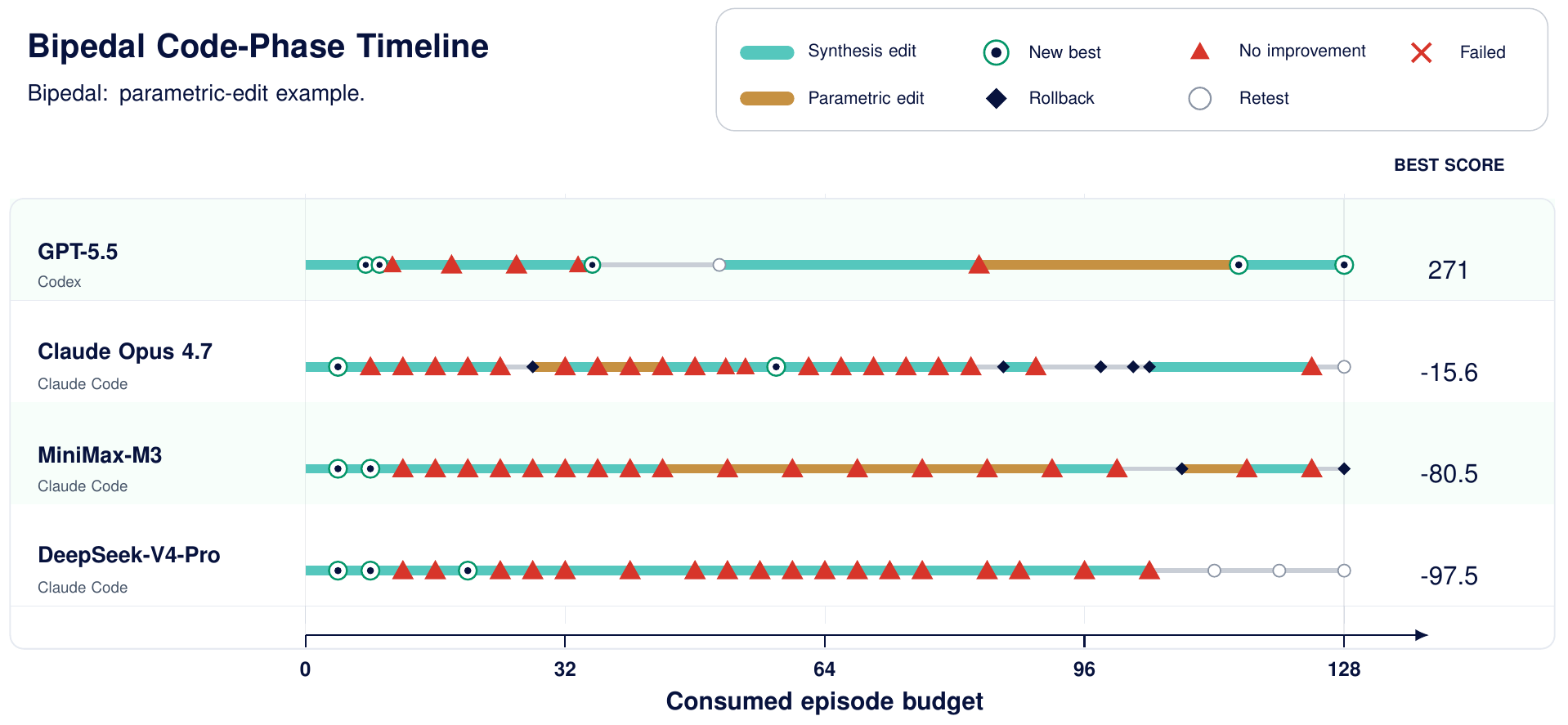}
    \caption{\small Bipedal code-phase timeline, rendered with the same synthesis-edit
    and parametric-edit phase rules as Figure~\ref{fig:carracing-code-phase}.
    The environment is tuning-dominant, but successful tuning still depends on
    first reaching a viable gait topology; same-topology source-bundle edits
    then expose whether an agent can improve that structure by adjusting
    constants and thresholds.}
    \label{fig:bipedal-code-phase}
\end{figure}

Aggregate edit counts show which transitions help, but not when agents invent
structure, tune it, or manage candidates. Figures~\ref{fig:carracing-code-phase}
and~\ref{fig:bipedal-code-phase} put the same edit classifier from
Table~\ref{tab:edit-types} back onto each run's budget axis. A timeline phase is
a run of adjacent score-bearing transitions with the same edit type: a
synthesis-edit phase begins when the numeric-constant-stripped source-bundle AST
topology changes, and a parametric-edit phase begins when the source bundle
changes while that topology is preserved. Rollbacks and retests are overlaid as
candidate-management events, not additional edit types.

The two timelines are randomly sampled case studies rather than an additional
selection based on which traces most cleanly support the aggregate split.
Before auditing phase histories, we sampled one environment from each demand
group: CarRacing from the synthesis-dominant group and BipedalWalker from the
tuning-dominant group. CarRacing makes the synthesis demand visible: agents
must turn pixels into a driving state, detect tracker failure, and preserve
recovery behavior across noisy visible rollouts. Figure~\ref{fig:carracing-code-phase} gives a directly countable phase pattern for the two high-return traces: Claude Opus 4.7 stays in synthesis-edit phases, while GPT-5.5 has one short
parametric-edit phase after early synthesis improvements before returning to
synthesis edits. BipedalWalker shows the complementary pattern. It is tuning-dominant, but tuning becomes useful only after a gait-producing topology exists. We operationalize that milestone by return: GPT-5.5 is the only run in this audit with a positive high-return gait, reaching timeline best score $271$ and validation-selected held-out return $248.874$ (Table~\ref{tab:core16-raw}); the other three BipedalWalker traces remain at negative timeline best scores ($-15.6$ or lower) and mostly churn structures or revisit candidates without crossing that return threshold.

%\runzhe{Explain why CarRacing and BipedalWalker were selected before interpreting them as complementary exemplars. If they were chosen after inspecting results, add at least a selection criterion. Replace ``successful runs therefore improve mostly'' with a directly countable statement from the timeline, and define ``stable gait'' using return or episode-completion evidence.}

% \begin{figure}[H]
%     \centering

%     \begin{minipage}[t]{0.495\textwidth}
%         \centering
%         \includegraphics[width=\textwidth]{figs/agent_audit_racing_timeline_1.pdf}
        
%         \vspace{0.2em}
%         \small \textbf{(a) CarRacing.} Synthesis-edit example.
%     \end{minipage}
%     \hfill
%     \begin{minipage}[t]{0.495\textwidth}
%         \centering
%         \includegraphics[width=\textwidth]{figs/agent_audit_bipedal_timeline_1.pdf}
        
%         \vspace{0.2em}
%         \small \textbf{(b) Bipedal.} Parametric-edit example.
%     \end{minipage}

%     \caption{\han{Code-phase timelines for CarRacing and Bipedal. Phase bands are inferred
%     mechanically from the same policy source-bundle rule as
%     Table~\ref{tab:edit-types}: synthesis-edit phases denote new AST topologies
%     after numeric constants are stripped, and parametric-edit phases denote
%     changed source bundles under the same topology. Symbols mark validation
%     outcomes and candidate-management events; rollback/retest are event types,
%     not additional edit types. Bipedal is tuning-dominant, but successful tuning
%     still depends on first reaching a viable gait topology; same-topology
%     source-bundle edits then expose whether an agent can improve that structure
%     by adjusting constants and thresholds.}}
%     \label{fig:code-phase-timelines}
% \end{figure}

\begin{figure}[t]
    \centering
    \includegraphics[width=0.98\textwidth]{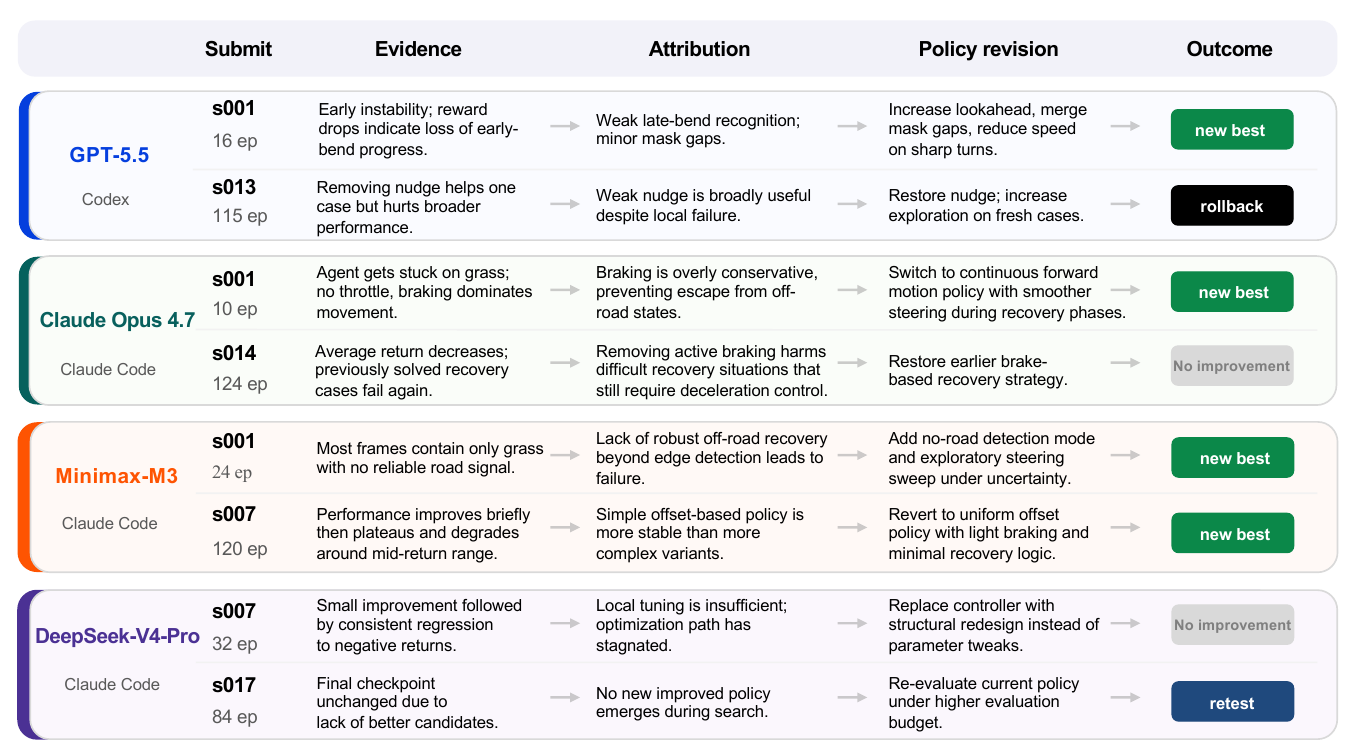}
    \caption{\small CarRacing feedback-utilization traces. Each row links evidence, attribution, policy revision, and outcome across agents. The submit column reports the submission index (s00k denotes the k-th submission) and the cumulative episode budget consumed prior to that submission. Labels are derived from logs, feedback summaries, and checkpoint diffs. The figure provides qualitative evidence of how feedback is translated into policy updates and is not an aggregate metric.
    % Audited CarRacing feedback-utilization trace. Each row links the
    % visible evidence an agent read, its failure attribution, the corresponding
    % code edit or candidate-selection action, and the observed submit outcome.
    % The Submit column reports the submit index and consumed episode budget; row
    % labels are audited from agent streams, feedback summaries, and checkpoint
    % diffs.
    % The figure is qualitative case evidence for how agents operationalize
    % feedback into policy revisions; it is not an aggregate metric and does not
    % imply hidden validation or held-out feedback was visible.
    }
    \label{fig:feedback-utilization-trace}
\end{figure}

Figure~\ref{fig:feedback-utilization-trace} unpacks the CarRacing timeline by connecting code phases to visible feedback. The successful traces do not merely react to scalar score changes: agents attribute visible failures to perception or control mechanisms, edit the corresponding policy structure, and use later feedback to select or roll back candidates. Weaker runs expose the same loop with less traction: mechanism replacements and retests occur, yet the controller rarely escapes the wrong abstraction. Appendix Figure~\ref{fig:gpt55-racing-diagnostics-panel} complements this timeline with the visible game frames that GPT-5.5 saved during testing, showing how it used rollout observations as diagnostic evidence for subsequent policy edits.

% \begin{figure}[H]
%     \centering
%     \includegraphics[width=0.98\textwidth]{figs/feedback_utilization_trace.pdf}
%     \caption{Audited CarRacing feedback-utilization trace. Each row links the
%     visible evidence an agent read, its failure attribution, the corresponding
%     code edit or candidate-selection action, and the observed submit outcome.
%     The Submit column reports the submit index and consumed episode budget; row
%     labels are audited from agent streams, feedback summaries, and checkpoint
%     diffs.
%     The figure is qualitative case evidence for how agents operationalize
%     feedback into policy revisions; it is not an aggregate metric and does not
%     imply hidden validation or held-out feedback was visible.}
%     \label{fig:feedback-utilization-trace}
% \end{figure}

Across these diagnostics, higher-scoring harness--model runs are associated with more successful structural edits on synthesis-dominant tasks and with qualitative traces that link visible failure evidence to targeted policy revisions.
%\runzhe{safer version is: ``Across these diagnostics, higher-scoring harness--model runs are associated with more successful structural edits on synthesis-dominant tasks and with qualitative traces that link visible failure evidence to targeted policy revisions.''}

\subsection{Limitations of the Diagnostics}
\label{sec:analysis-limits}

These diagnostics are conservative proxies, not semantic proofs. AST topology
captures objective code changes, but two topologies can implement similar
behavior, and one topology can mix useful and harmful ideas. The policy
source-bundle boundary includes \texttt{policy.py} and self-written helper
modules reachable from it, reducing a policy-only blind spot, but it still
excludes generated data files, learned weights, and unreferenced experiments.
The synthesis/tuning split is a lens rather than a taxonomy of tasks:
Bipedal still needs a viable gait structure before tuning helps, and CarRacing
still benefits from later parameter choices once perception and control are in
place. We therefore treat the figures as converging evidence from scores, code
artifacts, edit outcomes, and visible-feedback traces, not as calibrated
measurements of latent abilities.

\section{Conclusion}
\evopolicygym{} casts autonomous policy improvement as a controlled evaluation
of the systems agents build over time. Each harness--model agent edits an
executable policy under a fixed interaction budget, learns only from visible
train feedback, and is judged by hidden validation-selected heldout
performance. The Core16 results show that high scores require more than
isolated task wins: strong agents infer task-appropriate abstractions,
translate feedback into mechanism-level code changes, and preserve useful
candidates under budget pressure. By pairing leaderboard scores with
trajectory-level diagnostics, \evopolicygym{} provides a concrete protocol for
measuring stable, feedback-driven autonomous policy evolution.

\section*{Acknowledgements}
We thank Jiayi Weng for the public blog post \emph{Learning Beyond Gradients}
\citep{learningbeyondgradients}. Its central insight, that coding agents can
continually maintain and improve heuristic systems rather than merely produce
one-off policy files, directly shaped the starting point of this work. During
this project, we found that the word ``heuristic''
was difficult to make operational. Traditional hand-written rules are often
called heuristics, but the boundary becomes unclear once the policy includes
tuned numeric parameters, learned components, or baselines optimized by methods
such as PPO. This difficulty pushed us to make the policy system the benchmark
object: an executable policy together with the state, code structure, feedback
traces, and revision history that an agent can maintain. Fixed
environment-interaction budgets then create optimization pressure while hidden
validation and heldout splits prevent leakage. This setting lets us compare
how well stronger models extract insight from feedback and turn that insight
into policy improvements.

\bibliography{iclr2026_conference}
\bibliographystyle{iclr2026_conference}

\appendix

\newtcblisting{strategycode}{
  enhanced,
  listing only,
  breakable,
  width=\textwidth,
  colback=cardbg,
  colframe=accent,
  boxrule=0.35pt,
  arc=2pt,
  left=6pt,
  right=6pt,
  top=4pt,
  bottom=4pt,
  boxsep=0pt,
  listing options={
    language=Python,
    basicstyle=\ttfamily\scriptsize,
    keywordstyle=\color{accent}\bfseries,
    commentstyle=\color{black!55},
    stringstyle=\color{black!70},
    columns=fullflexible,
    keepspaces=true,
    showstringspaces=false,
    breaklines=true,
    breakatwhitespace=true
  }
}
\newpage

\section{Benchmark Object and Task Suite}
\label{app:benchmark-object}

\subsection{Run Protocol and Policy System}
\label{app:protocol-details}

An \evopolicygym{} run gives the agent a live workspace and a fixed interaction
budget for improving one executable policy system. The server stages
\texttt{AGENTS.md}, initializes \texttt{system/} with the policy entry point,
and exposes a local HTTP service. The agent uses \texttt{/info} to inspect the
remaining budget and protocol limits, \texttt{/task} to read the task and
policy contract, and \texttt{/submit} to request visible train rollouts.
Validation, held-out evaluation, and finalization remain server-side.

The policy entry point is \texttt{system/policy.py}. It exports a top-level
\texttt{Policy} class whose constructor receives the observation space, action
space, and environment metadata; the server then calls \texttt{reset} once per
episode and \texttt{act} at each environment step. Agents may add modules,
configuration files, tests, weights, memory files, and analysis utilities under
\texttt{system/}. For each submit, the server imports the submitted policy and
constructs a fresh \texttt{Policy} instance, so durable state lives in files
under \texttt{system/}.

Episode budget is charged by the expanded list of requested train case
indices. In Core16, each run uses a total budget of 128 episodes and allows
between 1 and 128 episodes per submit. The server preserves request order and
repeated indices. Request-format failures are rejected before snapshot and do
not consume budget. Once a request passes protocol validation, the server
snapshots \texttt{system/}; import errors, initialization errors, rollout
timeouts, and other execution-stage failures still consume the requested
episode budget and write visible feedback.

Feedback is written under \texttt{feedback/submit\_NNN/}. Each completed
submit has a \texttt{summary.json} containing status, requested case indices,
remaining budget, episode returns, episode lengths, errors, timing, and
aggregate return statistics. Successful episode directories contain step-level
\texttt{trajectory.jsonl} records, policy stdout/stderr streams, and optional
environment-specific media such as rendered video or external observation
arrays. Submit-level failures write \texttt{errors.txt}; per-episode failures
write the corresponding episode error file while preserving the rest of the
submit evidence when possible.

The workspace has live no-rollback semantics. The agent may overwrite its own
\texttt{system/} files after each submit, and harmful valid edits stay visible
until the agent repairs or restores them. The server separately stores
immutable submitted checkpoints for hidden validation and final held-out
evaluation.

\subsection{Core16 Suite}
\label{app:core16-suite-details}

Core16 selects 16 implemented Gymnasium-compatible scenarios from four
environment families. The suite spans control, visual driving, locomotion,
symbolic partial observability, manipulation, and traffic-style control,
requiring controllers, visual abstractions, world models, phase machines, and
recovery logic.

\begin{table}[H]
\caption{Implemented \evopolicygym{} scenarios used in the 128-episode
Core16 suite.}
\label{tab:suite}
\begin{center}
\scriptsize
\setlength{\tabcolsep}{3.5pt}
\resizebox{\textwidth}{!}{%
\begin{tabular}{llll}
\toprule
Category & Suite ID & Gymnasium env & Obs./action \\
\midrule
Gym / Box2D & \texttt{gym/acrobot} & \texttt{Acrobot-v1} & state/discrete \\
Gym / Box2D & \texttt{gym/continuouscar} & \texttt{MountainCarContinuous-v0} & state/continuous \\
Gym / Box2D & \texttt{gym/bipedal} & \texttt{BipedalWalker-v3} & state/continuous \\
Gym / Box2D & \texttt{gym/racing} & \texttt{CarRacing-v3} & image/continuous \\
MuJoCo & \texttt{gym/reacher5} & \texttt{Reacher-v5} & state/continuous \\
MuJoCo & \texttt{gym/halfcheetah5} & \texttt{HalfCheetah-v5} & state/continuous \\
MuJoCo & \texttt{gym/ant5} & \texttt{Ant-v5} & state/continuous \\
MuJoCo & \texttt{gym/pusher5} & \texttt{Pusher-v5} & state/continuous \\
MiniGrid & \texttt{gymnasium/MiniGrid-DoorKey-16x16-v0} & \texttt{MiniGrid-DoorKey-16x16-v0} & symbolic/discrete \\
MiniGrid & \texttt{gymnasium/MiniGrid-KeyCorridorS4R3-v0} & \texttt{MiniGrid-KeyCorridorS4R3-v0} & symbolic/discrete \\
MiniGrid & \texttt{gymnasium/MiniGrid-FourRooms-v0} & \texttt{MiniGrid-FourRooms-v0} & symbolic/discrete \\
MiniGrid & \texttt{gymnasium/MiniGrid-ObstructedMaze-1Q-v1} & \texttt{MiniGrid-ObstructedMaze-1Q-v1} & symbolic/discrete \\
Robotics / Driving & \texttt{gymnasium/parking-v0} & \texttt{parking-v0} & state/control \\
Robotics / Driving & \texttt{gymnasium/roundabout-v0} & \texttt{roundabout-v0} & state/control \\
Robotics / Driving & \texttt{gymnasium/FetchPush-v4} & \texttt{FetchPush-v4} & dict/continuous \\
Robotics / Driving & \texttt{gymnasium/FetchPickAndPlace-v4} & \texttt{FetchPickAndPlace-v4} & dict/continuous \\
\bottomrule
\end{tabular}
}
\end{center}
\end{table}

The four-by-four organization gives each family equal weight in category-level
analysis while preserving heterogeneous policy interfaces. State-control tasks
test compact feedback interpretation, CarRacing adds visual abstraction,
MiniGrid stresses persistent symbolic state, and the Fetch tasks require
geometric phase control. This mix is why raw returns are reported per
environment and aggregated only after within-environment ranking.

\section{Evaluation Protocol and Agent Configuration}
\label{app:evaluation-protocol}

\subsection{Visibility, Selection, and Scoring}
\label{app:evaluation-and-scoring}
\label{sec:hidden-eval}
\label{sec:scoring-audit}

Each Core16 run uses three disjoint case splits. Train cases are visible as
integer handles and provide all in-loop feedback. Validation and held-out cases
are server-side. For the experiments in this paper, hidden validation contains
16 cases and hidden held-out evaluation contains 32 cases per environment. The
agent never sees validation or held-out case identities, trajectories, returns,
or failure details during optimization.

\begin{table}[H]
\caption{Visibility boundary for agent-facing artifacts.}
\label{tab:visibility}
\begin{center}
\begin{tabular}{lll}
\toprule
Split & Agent-visible during run & Server role \\
\midrule
Train & IDs, summaries, trajectories, failures & Online revision signal \\
Validation & None & Private checkpoint selection \\
Held-out & None & Final generalization measurement \\
\bottomrule
\end{tabular}
\end{center}
\end{table}

After the 128-episode train budget is exhausted, the server evaluates every
\texttt{status == ok} checkpoint on the hidden validation split. The checkpoint
with the highest validation mean return is selected; equal validation means are
resolved by choosing the later submit. The selected checkpoint is then
evaluated on the held-out split. The raw held-out mean return is the
per-environment number reported in Table~\ref{tab:core16-raw}.

Cross-environment aggregation uses only ranks within each environment. Let
$y_{m,e}$ be the held-out mean return for entry $m$ on environment $e$, and
let $\mathrm{rank}_e(m)$ be the descending rank of $y_{m,e}$ among the four
reported agents plus the uniform random-policy reference. For agents,
$y_{m,e}$ is validation-selected; for the random-policy reference, it is the
held-out mean return of uniform random actions on the same 32 held-out cases,
with no training budget or validation selection. Equal held-out means share the
same rank score. The per-environment score is
\[
s_{m,e}=1-\frac{\mathrm{rank}_e(m)-1}{N_e-1},
\]
where $N_e=5$ in the Core16 leaderboard. Category scores average $s_{m,e}$
over the four environments in that category. The Core16 score averages
$s_{m,e}$ over all 16 environments. These aggregate scores are analysis
metrics; hidden validation selection and held-out evaluation operate on raw
environment returns.

This scoring separates two questions. Raw held-out means show how strong a
selected policy is on one environment. Rank-normalized scores summarize which
agent more consistently converts the same interaction budget into competitive
policies across reward scales.

Runs also record audit signals: accepted and rejected submits, budget consumed
per submit, post-hoc validation best-so-far curves, selected checkpoint,
invalid-transition rate, score-drop events, policy complexity growth, wall
time, and agent token accounting when available. These traces support the
behavior analysis in Section~\ref{sec:behavior-analysis}.

\subsection{Agent and Run Configuration}
\label{app:agent-run-configuration}

All four leaderboard agents use the same run-level configuration: the Core16
environment list, a 128-episode train interaction budget, minimum submit size
1, maximum submit size 128, hidden validation size 16, and hidden held-out size
32. Each run uses external train/validation/held-out case files for the
corresponding environment. The server binds to a local loopback address with an
ephemeral port, and the port is exposed to the agent only through the staged
run instructions. The random-policy reference in Tables~\ref{tab:core16-raw}
and~\ref{tab:aggregate} is evaluated directly on the same held-out cases and
does not use this agent harness configuration.

\begin{table}[H]
\caption{Harness configuration for the Core16 leaderboard agents.}
\label{tab:agent-config}
\begin{center}
\footnotesize
\setlength{\tabcolsep}{4pt}
\begin{tabular}{llll}
\toprule
Agent & Harness & Model string & Shared settings \\
\midrule
GPT-5.5 & Codex & \texttt{gpt-5.5} & retries 5, backoff 4.0, bypass enabled \\
Claude Opus 4.7 & Claude Code & \texttt{claude-opus-4-7-thinking-max} & retries 5, backoff 4.0 \\
MiniMax-M3 & Claude Code & \texttt{MiniMax-M3} & retries 5, backoff 4.0 \\
DeepSeek-V4-Pro & Claude Code & \texttt{deepseek-v4-pro} & retries 5, backoff 4.0 \\
\bottomrule
\end{tabular}
\end{center}
\end{table}

The Claude Code-compatible runs expose the same tool profile:
\texttt{Bash}, \texttt{Read}, \texttt{Edit}, \texttt{Write}, \texttt{Glob}, and
\texttt{Grep}, with bypass-style permissions. The Codex run uses the Codex
adapter with a persistent logical session. Retries handle harness or service
timeouts and exceptions; retry events do not add environment interaction and do
not change the server-side budget accounting.

The shared run budget fixes the environment-interaction comparison. Harness
differences still affect context management and tool traces, so token and cost
statistics are diagnostic only and excluded from scoring.

\section{Supplementary Analysis Diagnostics}
\label{app:additional-diagnostics}

This section expands the quantitative diagnostics used in
Section~\ref{sec:behavior-analysis}. The ability table provides the
per-environment values behind the synthesis/tuning split; the token table
describes harness-level context traffic; and the edit-size plot summarizes how
large checkpoint changes relate to visible improvement. These diagnostics
explain behavior but do not affect validation selection, held-out evaluation, or
leaderboard rank.

\subsection{Per-Environment Relative Held-Out Performance by Task Demand}
\label{app:ability-split-detail}

\begin{table}[H]
\caption{Per-environment values behind Figure~\ref{fig:two-abilities}. Scores
are scaled between a random policy ($0$) and the best evaluated agent
performance on that environment ($1$), with random anchors measured on the
same held-out pools.}
\label{tab:two-abilities-detail}
\begin{center}
\footnotesize
\setlength{\tabcolsep}{5.5pt}
\begin{tabular}{@{}lrrrr@{}}
\toprule
Demand / environment & GPT-5.5 & Claude Opus 4.7 & MiniMax-M3 & DeepSeek-V4-Pro \\
\midrule
\multicolumn{5}{@{}l}{\emph{Synthesis-dominant}} \\
\quad DoorKey & 1.00 & 1.00 & 0.00 & 0.00 \\
\quad KeyCorridor & 0.99 & 1.00 & 0.00 & 0.00 \\
\quad ObstructedMaze & 0.99 & 1.00 & 0.00 & 0.00 \\
\quad FourRooms & 0.94 & 1.00 & 0.55 & 0.06 \\
\quad CarRacing & 1.00 & 1.00 & 0.42 & 0.09 \\
\textbf{Synthesis mean} & \textbf{0.98} & \textbf{1.00} & \textbf{0.19} & \textbf{0.03} \\
\midrule
\multicolumn{5}{@{}l}{\emph{Tuning-dominant}} \\
\quad Parking & 1.00 & 0.54 & 0.86 & 0.00 \\
\quad Bipedal & 1.00 & 0.24 & 0.06 & 0.01 \\
\quad HalfCheetah & 1.00 & 0.83 & 1.00 & 0.32 \\
\quad FetchPush & 1.00 & 0.95 & 0.72 & 0.66 \\
\quad Roundabout & 0.87 & 0.71 & 0.77 & 1.00 \\
\quad FetchPickAndPlace & 1.00 & 0.95 & 0.96 & 0.73 \\
\quad Pusher & 1.00 & 0.99 & 0.98 & 0.85 \\
\quad Acrobot & 1.00 & 0.99 & 0.88 & 0.98 \\
\quad Ant & 1.00 & 1.00 & 0.99 & 0.91 \\
\quad Reacher & 1.00 & 0.99 & 0.96 & 0.92 \\
\quad ContinuousCar & 0.97 & 1.00 & 0.95 & 0.97 \\
\textbf{Tuning mean} & \textbf{0.99} & \textbf{0.83} & \textbf{0.83} & \textbf{0.67} \\
\bottomrule
\end{tabular}
\end{center}
\end{table}

\subsection{Token and Cost Accounting}
\label{app:cost-diagnostics}
\label{app:token-accounting}

Token accounting is diagnostic and excluded from the leaderboard score. We
separate non-cached input, cache read/creation, and output tokens because cache
events overlap semantically with previously supplied context. Values are parsed
from agent stream logs and reported in millions. For Codex streams, non-cached
input subtracts \texttt{cached\_input\_tokens} from \texttt{input\_tokens};
Claude Code streams already report cache read and cache creation separately.

\begin{table}[t]
\caption{Per-task agent stream token accounting for the \texttt{main-128}
Core16 runs. Values are parsed from agent stream logs and reported in millions;
model subcolumns are non-cached input (In), cache read plus cache creation
(Cache), and output including Codex reasoning output (Out). For Codex streams,
In subtracts \texttt{cached\_input\_tokens} from \texttt{input\_tokens};
Claude Code streams already report cache read/creation in separate fields.
Dashes mark missing stream logs. These diagnostic columns are not used in the
leaderboard rank score.}
\label{tab:token-accounting}
\begin{center}
\scriptsize
\setlength{\tabcolsep}{1.8pt}
\textbf{A. Gym / Box2D and MuJoCo}\par\vspace{0.25em}
\begin{tabular}{@{}l*{4}{rrr}@{}}
\toprule
 & \multicolumn{3}{c}{GPT-5.5} & \multicolumn{3}{c}{Claude Opus 4.7}
 & \multicolumn{3}{c}{MiniMax-M3} & \multicolumn{3}{c}{DeepSeek-V4-Pro} \\
\cmidrule(lr){2-4}\cmidrule(lr){5-7}\cmidrule(lr){8-10}\cmidrule(lr){11-13}
Task & In & Cache & Out & In & Cache & Out & In & Cache & Out & In & Cache & Out \\
\midrule
Acrobot & 0.06 & 0.61 & 0.03 & \textless{}0.01 & 5.85 & 0.06 & 0.25 & 6.05 & 0.09 & 0.06 & 2.48 & 0.04 \\
Cont.Car & 0.10 & 1.22 & 0.02 & \textless{}0.01 & 7.07 & 0.10 & 0.04 & 2.92 & 0.06 & 0.07 & 3.62 & 0.05 \\
Bipedal & 0.19 & 2.67 & 0.05 & \textless{}0.01 & 13.85 & 0.11 & 0.31 & 9.78 & 0.11 & 0.06 & 6.24 & 0.08 \\
CarRacing & 0.53 & 11.36 & 0.09 & \textless{}0.01 & 28.90 & 0.12 & 0.73 & 10.80 & 0.06 & 0.37 & 44.19 & 0.27 \\
Reacher & 0.18 & 2.08 & 0.03 & \textless{}0.01 & 27.07 & 0.17 & 0.16 & 7.29 & 0.09 & 0.05 & 2.78 & 0.07 \\
HalfCheetah & 0.08 & 0.78 & 0.03 & \textless{}0.01 & 6.50 & 0.08 & 0.22 & 12.69 & 0.08 & 0.07 & 5.32 & 0.06 \\
Ant & 0.11 & 2.26 & 0.05 & \textless{}0.01 & 6.66 & 0.08 & 0.13 & 7.03 & 0.06 & 0.06 & 4.74 & 0.07 \\
Pusher & 0.13 & 1.79 & 0.04 & \textless{}0.01 & 9.25 & 0.13 & 0.29 & 7.44 & 0.12 & 0.06 & 3.30 & 0.07 \\
\bottomrule
\end{tabular}

\vspace{0.75em}
\textbf{B. MiniGrid and Robotics / Driving}\par\vspace{0.25em}
\begin{tabular}{@{}l*{4}{rrr}@{}}
\toprule
 & \multicolumn{3}{c}{GPT-5.5} & \multicolumn{3}{c}{Claude Opus 4.7}
 & \multicolumn{3}{c}{MiniMax-M3} & \multicolumn{3}{c}{DeepSeek-V4-Pro} \\
\cmidrule(lr){2-4}\cmidrule(lr){5-7}\cmidrule(lr){8-10}\cmidrule(lr){11-13}
Task & In & Cache & Out & In & Cache & Out & In & Cache & Out & In & Cache & Out \\
\midrule
DoorKey & 0.14 & 1.59 & 0.04 & \textless{}0.01 & 3.80 & 0.08 & 0.51 & 12.54 & 0.28 & 0.17 & 16.73 & 0.27 \\
KeyCorr. & 0.07 & 1.05 & 0.04 & 0.01 & 7.09 & 0.09 & 0.96 & 14.03 & 0.20 & 0.16 & 13.08 & 0.29 \\
FourRooms & 0.23 & 3.36 & 0.07 & \textless{}0.01 & 12.08 & 0.17 & 0.34 & 9.55 & 0.11 & 0.25 & 22.02 & 0.35 \\
Obst.Maze & 0.39 & 12.27 & 0.12 & \textless{}0.01 & 34.88 & 0.24 & 0.35 & 19.60 & 0.18 & 0.24 & 21.99 & 0.38 \\
Parking & 0.17 & 3.09 & 0.05 & \textless{}0.01 & 0.35 & \textless{}0.01 & 0.92 & 17.46 & 0.13 & 0.12 & 9.57 & 0.13 \\
Roundabout & 0.25 & 2.23 & 0.04 & \textless{}0.01 & 7.48 & 0.06 & 0.12 & 4.52 & 0.05 & 0.07 & 2.90 & 0.06 \\
FetchPush & 0.22 & 3.49 & 0.05 & \textless{}0.01 & 9.05 & 0.15 & 0.24 & 11.90 & 0.12 & 0.06 & 4.37 & 0.08 \\
FetchPick & 0.10 & 2.06 & 0.04 & \textless{}0.01 & 8.05 & 0.10 & 0.37 & 6.72 & 0.13 & 0.09 & 4.70 & 0.07 \\
\bottomrule
\end{tabular}
\end{center}
\end{table}

The table shows that token use varies substantially across tasks and harnesses.
Cache traffic can dominate non-cached input in several runs, especially when the
agent carries long interaction histories across repeated revisions. These
numbers help interpret optimization behavior and implementation overhead, while
the leaderboard remains tied to fixed environment interaction.

\subsection{Edit-Size Diagnostic}
\label{app:edit-size-diagnostic}

\begin{figure}
    \centering
    \includegraphics[width=0.78\textwidth]{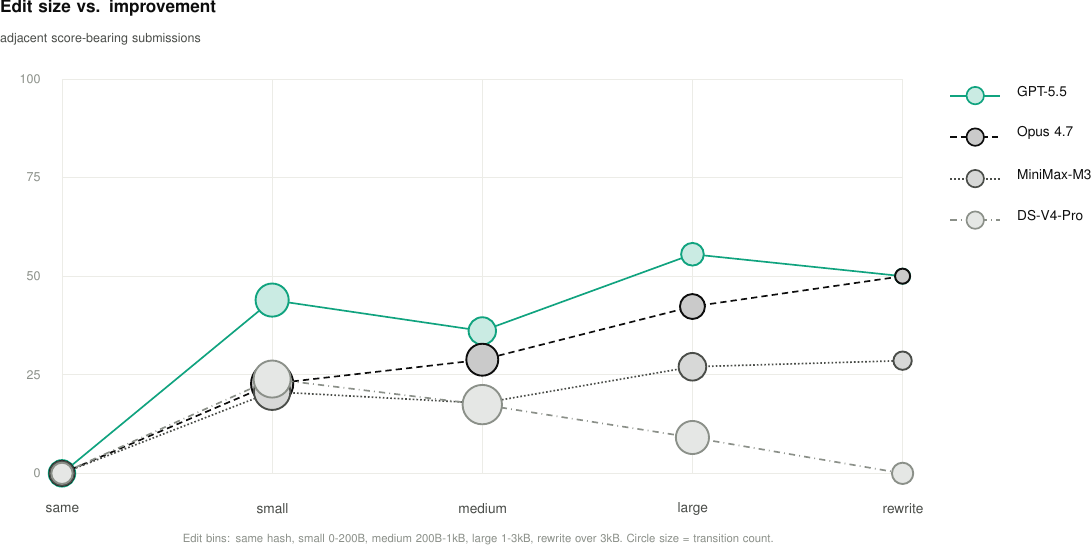}
    \caption{Observed association between policy edit size and improvement
    probability across adjacent submissions. Edit bins are computed from
    checkpoint code diffs; circle size indicates the number of transitions in
    each bin. The plot is diagnostic: large edits can represent useful
    mechanism synthesis, destructive rewrites, or
    interface repair depending on the surrounding run context.}
    \label{fig:code-edit-risk-return}
\end{figure}

Same-hash transitions have zero improvement by construction, and small or
medium edits form the common local-search regime. GPT-5.5 and Claude Opus 4.7
still improve on a meaningful share of larger structural edits, which matches
their qualitative traces: both agents introduce mechanisms and then consolidate
them. MiniMax-M3 obtains some large gains from rewrites but consolidates them
less reliably, while DeepSeek-V4-Pro less often turns large edits into
validation-best checkpoints.

\section{Policy Mechanism Case Studies}
\label{app:case-study-strategies}

\begin{figure}[H]
    \centering
    \includegraphics[width=0.98\textwidth]{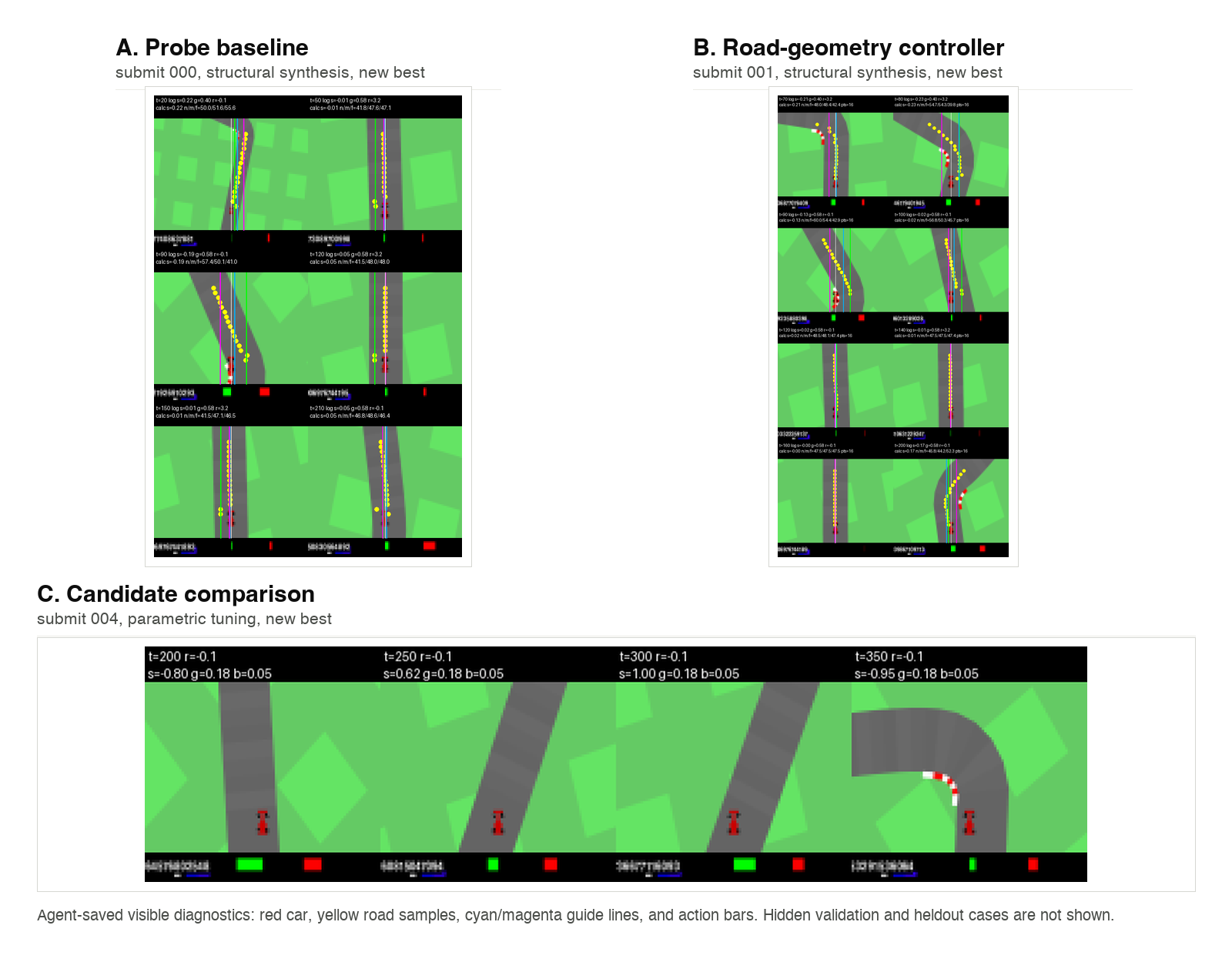}
    \caption{GPT-5.5 CarRacing visible diagnostics saved by the agent during
    the run. Panels A--B show how structural-synthesis edits turn pixel
    observations into road-geometry control signals: yellow points mark sampled
    road evidence, cyan/magenta lines mark guide estimates, and action bars/log
    text summarize the agent's own rollout diagnostics. Panel C shows a later
    visible candidate comparison after a parametric-tuning edit. These images
    are qualitative evidence only and do not expose hidden-validation or
    held-out cases.}
    \label{fig:gpt55-racing-diagnostics-panel}
\end{figure}

The quantitative diagnostics above describe aggregate behavior; this section
shows what successful submitted policies look like. Figure~\ref{fig:gpt55-racing-diagnostics-panel}
grounds the CarRacing trace in agent-visible diagnostic artifacts. The snippets
distill representative submitted policy mechanisms from checkpoint code and
nearby stream evidence, keeping the causal structure of each strategy while
omitting interface boilerplate, constants, and environment wrappers.

\paragraph{CarRacing: road-mask lookahead and recovery.}
The strongest CarRacing policies first convert pixels into a road mask, then
trace near/mid/far centers, combine lookahead curvature with edge warnings,
and reduce speed when visual confidence drops. This turns raw frames into a
closed-loop controller with an explicit recovery mode.
\begin{strategycode}
def act(self, obs):
    centers = trace_road_centers(road_mask(obs), last_target=self.last_target)
    if not centers:
        self.lost_steps += 1
        center = global_road_center(obs)
        if center is None:
            steer = recover_toward_last_track(self.last_steer, self.last_target)
        else:
            raw = clip((center - self.center_x) / 23.0, -1.0, 1.0)
            steer = 0.45 * self.last_steer + 0.55 * raw
        self.last_steer = clip(steer, -1.0, 1.0)
        return [float(self.last_steer), 0.25, 0.0]

    near, mid, far = summarize_centers(centers)
    curvature = (far - mid) / self.width
    raw_steer = 0.75 * dev(mid) + 1.20 * dev(far) + 0.60 * curvature
    raw_steer = repair_with_global_center(raw_steer, obs, edge_alert=True)
    steer = clip(0.45 * self.last_steer + 0.55 * raw_steer, -1.0, 1.0)
    self.last_steer = steer
    self.last_target = 0.25 * near + 0.40 * mid + 0.35 * far

    turn = max(abs(steer), abs(raw_steer), abs(dev(far)))
    gas, brake = speed_schedule(turn)
    return [float(steer), float(gas), float(brake)]
\end{strategycode}

\paragraph{HalfCheetah: periodic gait with safety scaling.}
HalfCheetah exposes open-loop synthesis: agents search for a compact
oscillatory gait, then wrap it with clipping and posture-based amplitude
scaling. The policy system stores the gait parameters, and visible returns
tune phase, amplitude, and frequency.
\begin{strategycode}
def act(self, obs):
    root_height = float(obs[ROOTY])
    excess = max(0.0, abs(root_height) - self.safety_thresh)
    scale = max(self.safety_floor, 1.0 - excess)

    phase = 2.0 * math.pi * self.freq * self.t * self.dt
    hip = self.hip_amp * math.sin(phase)
    knee = self.knee_amp * math.sin(phase + math.pi / 2.0)
    ankle = self.ankle_amp * math.sin(phase + math.pi)
    action = np.array([
        hip,
        knee,
        ankle,
        -0.8 * hip,
        -0.7 * knee,
        -0.5 * ankle,
    ], dtype=np.float32)

    self.t += 1
    return np.clip(scale * action, self.action_low, self.action_high)
\end{strategycode}

\paragraph{ObstructedMaze: egocentric mapping and BFS planning.}
Successful MiniGrid policies build a persistent symbolic world model from the
7-by-7 egocentric image, update pose using previous actions, and plan toward
task objects with key, door, and blocker state. The same planner supports
frontier exploration, door toggling, pickup, drop, and obstacle clearing.
\begin{strategycode}
def act(self, obs):
    image = obs["image"]
    self.sync_pose(obs["direction"], self.last_action, self.last_front)
    self.update_world_from_view(image, pose=self.pose)
    front = self.cell_in_front(image)

    if self.target_ball_visible(front):
        action = PICKUP
    elif self.carrying_is_obstacle():
        action = self.drop_at_safe_cell(front)
    elif self.carrying_is_key() and self.door_blockers(self.key_color):
        action = self.clear_blocker_before_door(front)
    else:
        targets = (self.ball_targets() or self.key_targets()
                   or self.door_targets() or self.blocker_targets()
                   or self.frontier_targets())
        path = bfs(self.world, self.pose, targets, can_enter=self.can_enter)
        action = self.follow(path) if path else self.unstick_action()

    self.last_action = action
    self.last_front = front
    return action
\end{strategycode}

\paragraph{FetchPush: geometry-based phase controller.}
FetchPush policies compute the push direction from object to goal, move the
gripper behind the object, lower to a pushing height, and drive through the
object toward a point beyond the goal. Later repairs add a clearance phase for
cases where the gripper starts on the wrong side of the object.
\begin{strategycode}
def act(self, obs):
    raw = obs["observation"]
    grip = raw[0:3]
    obj = obs.get("achieved_goal", raw[3:6])
    goal = obs["desired_goal"]

    direction = unit(goal[:2] - obj[:2])
    distance_xy = norm(goal[:2] - obj[:2])
    behind = obj[:2] - 0.065 * direction
    along = dot(grip[:2] - obj[:2], direction)
    lateral = cross2(grip[:2] - obj[:2], direction)

    push_z = obj[2] + 0.018
    approach_z = obj[2] + 0.075
    clear_z = obj[2] + 0.115
    needs_clear = distance_xy < 0.130 and abs(lateral) < 0.045 and along > -0.020

    if distance_xy < 0.050:
        target, gains = [grip[0], grip[1], obj[2] + 0.095], (0.0, 12.0)
    elif needs_clear and grip[2] < clear_z - 0.025:
        target, gains = [grip[0], grip[1], clear_z], (0.0, 14.0)
    elif norm(behind - grip[:2]) > 0.030:
        target, gains = [behind[0], behind[1], approach_z], (20.0, 12.0)
    elif abs(grip[2] - push_z) > 0.022:
        target, gains = [behind[0], behind[1], push_z], (16.0, 16.0)
    else:
        target = [goal[0] + 0.075 * direction[0],
                  goal[1] + 0.075 * direction[1], push_z]
        gains = (22.0, 12.0)

    return servo_to_target(grip, target, xy_gain=gains[0], z_gain=gains[1])
\end{strategycode}

Across the four examples, the useful policies are small stateful programs:
they build a task abstraction, attach a controller or planner to it, and add
recovery logic for the failure modes exposed by visible feedback. This pattern
is the mechanism-level counterpart of the synthesis and tuning behavior in the
main analysis.

\end{document}